\documentclass[10pt,twocolumn,letterpaper]{article}

\usepackage[pagenumbers]{wacv} % To force page numbers, e.g. for an arXiv version

\usepackage{amsmath}

\usepackage{mathtools}

\usepackage{animate}

\usepackage{float}  
\usepackage{enumitem}

\usepackage{todonotes}
\usepackage{multirow}
\usetikzlibrary{arrows.meta}
\usetikzlibrary{backgrounds}

\usepackage{bbm}

\usepackage{pifont}
\usepackage{algorithm}
\usepackage{algorithmic}

\usepackage{stfloats}

\def\ourLoss{QAL}

\def\Cov{Cov}

\def\cSP{\overline{SP}}

\definecolor{wacvblue}{rgb}{0.21,0.49,0.74}
\usepackage[pagebackref,breaklinks,colorlinks,allcolors=wacvblue]{hyperref}

\def\wacvPaperID{2570} % *** Enter the WACV Paper ID here
\def\confName{WACV}
\def\confYear{2026}

\title{\ourLoss : A Loss for Recall–Precision Balance in 3D Reconstruction}

\author{
Pranay Meshram  \quad
Yash Turkar \quad
Kartikeya Singh \quad
Praveen Raj Masilamani  \quad \\
Charuvahan Adhivarahan \quad
Karthik Dantu\\[2mm]
University at Buffalo, The State University of New York, Buffalo, NY, USA\\
{\tt\small \{pranaywa, yashturk, ksingh35, pmasilam, charuvah, kdantu\}@buffalo.edu}
}

\begin{document}
\maketitle

\begin{abstract}
% Volumetric learning underpins many 3D vision tasks such as completion, reconstruction, and mesh generation, yet training still relies on Chamfer Distance (CD) or Earth Mover’s Distance (EMD), which fail to balance recall and precision. We introduce Quality-Aware Loss (QAL), a drop-in replacement for CD/EMD that combines (i) a coverage-weighted nearest-neighbor term with a soft $\epsilon$-margin to emphasize under-represented regions, and (ii) an uncovered--ground-truth attraction term that explicitly drives recall. 
% Across diverse pipelines, QAL delivers robust Coverage improvements, averaging +4.3\% over CD and +2.8\% over the best alternatives. Extensive ablations confirm stable performance across hyperparameters, and evaluations on PCN and ShapeNet benchmarks show QAL’s consistent generalization across datasets and architectures. QAL-trained completions achieve higher grasp scores than baselines under GraspNet evaluation, yielding more reliable downstream performance for robotic grasping. QAL thus provides a practical, plug-and-play objective for robust 3D vision and robotics pipelines.

Volumetric learning underpins many 3D vision tasks such as completion, reconstruction, and mesh generation, yet training objectives still rely on Chamfer Distance (CD) or Earth Mover’s Distance (EMD), which fail to balance recall and precision. We propose Quality-Aware Loss (QAL), a drop-in replacement for CD/EMD that combines a coverage-weighted nearest-neighbor term with an uncovered--ground-truth attraction term, explicitly decoupling recall and precision into tunable components. 

Across diverse pipelines, QAL achieves consistent coverage gains, improving by an average of +4.3 pts over CD and +2.8 pts over the best alternatives. Though modest in percentage, these improvements reliably recover thin structures and under-represented regions that CD/EMD overlook. Extensive ablations confirm stable performance across hyperparameters and across output resolutions, while full retraining on PCN and ShapeNet demonstrates generalization across datasets and backbones. Moreover, QAL-trained completions yield higher grasp scores under GraspNet evaluation, showing that improved coverage translates directly into more reliable robotic manipulation. 

QAL thus offers a principled, interpretable, and practical objective for robust 3D vision and safety-critical robotics pipelines.

\end{abstract}

\section{Introduction}
\label{sec:intro}
\begin{figure*}[!ht]
  \centering
  \vspace{-0.6cm}
    \includegraphics[width=\textwidth]{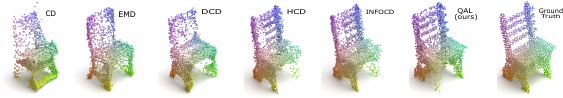}
   \vspace{-0.6cm}
    \caption{Point cloud completion with ECG~\cite{pan_ecg_2020} on \textbf{MVP}~\cite{pan2021variational}; Input partial clouds are omitted for space; all methods use identical inputs.} 
    %\ourLoss{} outperforms CD, EMD, DCD, HCD, and InfoCD. Input partial clouds are omitted for space; all methods use identical inputs.}  
  \label{fig:intro}
  \vspace{-0.6cm}
\end{figure*}
3D perception has become central to applications such as autonomous navigation, infrastructure inspection, dexterous manipulation, and object and scene recognition. This trend is reflected in a growing body of algorithms for 3D object detection and recognition~\cite{qiDeepHoughDetection2019,Zhou_2018_CVPR,Shi_2019_CVPR,Yang_2018_CVPR,Lang_2019_CVPR}, scene segmentation and classification~\cite{qiPointNetDeepHierarchical2017,Zhao_2021_ICCV,lai2022stratified,xu2020squeezesegv3}, and point cloud completion and estimation~\cite{yuan_pcn_2018,tchapmi_topnet_2019,pan_ecg_2020,pan2021variational,Huang_2020_CVPR}. Several factors have enabled this shift from 2D to 3D point-cloud representation: (i) advances in GPUs and mobile hardware~\cite{pandey2022GPU}, (ii) widespread availability of depth sensors and LiDAR~\cite{li2022LidarSurvey}, and (iii) deep learning architectures that can capture fine spatial variations and generalize from limited training data. Together, these developments have made large-scale 3D learning feasible and impactful. 

% Despite this progress, most pipelines are still trained with Euclidean distance-based losses such as \emph{Chamfer Distance (CD)} and \emph{Earth Mover’s Distance (EMD)}. Although effective for optimization, these objectives are fundamentally one-dimensional: they measure average point distances but ignore the recall-precision trade-off critical to 3D tasks. As a result, CD/EMD often fail to balance recall and precision: missing thin or occluded structures create holes, while over-dense predictions form spurious clusters. Both degrade downstream tasks such as grasping and navigation.
% \kar{upfront, we need to better clarify the motivation. What is recall here? What is precision here? What is wrong with a 1-D measure? - maybe call it scalar measure. It might be useful to state the motivation with an example. }

% As illustrated in \cref{fig:intro}, CD and its variants tolerate holes and spurious points (SP), often missing thin or occluded structures. By contrast, QAL increases coverage while controlling SP, enabling reconstructions that capture finer details of object geometry. Unlike HCD, which embeds distances in hyperbolic space, or InfoCD, which adds an information-theoretic margin, QAL directly encodes the recall–precision trade-off, making it tunable and aligned with downstream metrics.
% \kar{I think you are trying to do what I want - but I think we can do this better. Lets chat early morning Thur to see if we can converge on these two paragraphs. }

3D perception plays a critical role in applications such as autonomous navigation\cite{Hu_2023_CVPR}, dexterous manipulation\cite{Wan_2023_ICCV}, and scene reconstruction\cite{Chu_2023_CVPR}, where reliable geometric understanding is essential. Consider the task of \textbf{point cloud completion}~\cite{yuan_pcn_2018,pan_ecg_2020}, where the goal is to recover missing 3D geometry from partial observations. As shown in Fig.~\ref{fig:intro}, when an ECG~\cite{pan_ecg_2020} model is trained with Chamfer Distance (CD), thin or fine structures are often missing (low recall), while other regions exhibit spurious artifacts such as over-dense clusters (low precision). This illustrates a broader issue: although widely used, Earth mover's distance (EMD) and CD are \textit{scalar distance measures}---they average point-to-point errors but fail to capture nuances of the 3D comparisons such as coverage across the whole scene (precision) or the presence/absence of spurious artifacts (recall). In particular, CD tolerates holes and clustering, while EMD enforces one-to-one matching that reduces clustering but often leads to over-smoothed completions where fine details are blurred. Moreover, EMD is computationally expensive, making it impractical at scale and seldom adopted in recent pipelines. Both limitations degrade downstream tasks such as robotic grasping and navigation.

% While several variants of CD have been proposed to alleviate these issues, they still treat recall and precision implicitly rather than encoding their balance. 
% As illustrated in Fig.~\ref{fig:intro}, even recent variants of CD fail to adequately recover thin structures or suppress spurious artifacts, underscoring the need for a more principled approach. 
% Unlike prior CD variants that only reweight distances, QAL explicitly decouples recall and precision into two complementary terms, making the trade-off tunable and interpretable. 
% In this work, we introduce \textbf{Quality-Aware Loss (QAL)}, which directly formulates the recall–precision trade-off, increasing coverage of missing regions while controlling spurious points.

% To address these shortcomings, we introduce \textbf{Quality-Aware Loss (QAL)}, a differentiable recall--precision objective that increases coverage while controlling over-prediction. QAL is a \emph{drop-in replacement} for CD/EMD that aligns training with thresholded metrics such as Coverage and F1. Across diverse pipelines—including point-cloud completion, single-view 3D reconstruction, and image-to-mesh generation—QAL consistently improves coverage, with a modest rise in Spurious Points (SP) characteristic of recall-oriented optimization; \textbf{CD and F1 vary by backbone and dataset and are often comparable or improved}. This recall-prioritizing control is valuable for safety-critical robotics, where missing geometry is more costly than mild over-prediction.

While several variants of CD have been proposed to alleviate these issues, they still treat recall and precision implicitly rather than encoding their balance. 
Because partial observations are sparse and under-constrained, recovering missing regions (high recall) inevitably increases the risk of spurious predictions (lower precision), making explicit control of this trade-off essential. 
As illustrated in Fig.~\ref{fig:intro}, even recent variants of CD fail to adequately recover thin structures or suppress spurious artifacts, underscoring the need for a more principled approach. 
We introduce \textbf{Quality-Aware Loss (QAL)}, a differentiable recall--precision objective that explicitly decouples coverage and attraction terms, making the trade-off tunable and interpretable. 
QAL acts as a \emph{drop-in replacement} in any setting where CD, EMD, or their variants are used, aligning optimization with thresholded metrics such as Coverage and F1. 
Across diverse pipelines—including point-cloud completion, single-view 3D reconstruction, and image-to-mesh generation—QAL consistently improves coverage, with a modest rise in Spurious Points (SP) characteristic of recall-oriented optimization; CD and F1 vary by backbone and dataset but are often comparable or improved. 
This recall-prioritizing control is particularly valuable for safety-critical robotics tasks, where missing geometry is more costly than mild over-prediction.

The contributions of this work are:
\begin{itemize}
  \item \textbf{Quality-Aware Loss (QAL).} We introduce QAL, a drop-in replacement for Chamfer and Earth Mover’s Distance that explicitly balances recall and precision by combining coverage-weighted matching with an uncovered--GT attraction term.  
  \item \textbf{Comprehensive evaluation.} QAL is validated across completion, single-view reconstruction, and mesh generation, and further shown to improve grasp success when integrated into a pre-trained GraspNet model.  
  \item \textbf{Practical impact.} We align training with thresholded metrics (Cov@$\,\epsilon$, SP@$\,\epsilon$) and release efficient GPU implementations and code to enable adoption in 3D vision pipelines.  
\end{itemize}

\section{Related Work}

\noindent\textbf{Point-set distances and losses.}
Chamfer Distance (CD) and Earth Mover’s Distance (EMD) remain the most widely used objectives for training and evaluating point sets~\cite{fan2017point,achlioptas2018learning}. 
CD computes bidirectional nearest-neighbor (NN) discrepancies. It is fast and differentiable but sensitive to outliers, tolerates ``holes'' (low recall), and encourages clustered predictions that miss thin structures. 
EMD solves a one-to-one assignment (optimal transport) with informative gradients but is computationally heavy (exact solvers scale superlinearly; entropic/Sinkhorn approximations trade bias for speed). 
Several variants attempt to address these limitations: DCD reweights by local NN density to reduce over-/under-sampling~\cite{wu_density-aware_2021}; HCD embeds NN distances in hyperbolic space to preserve fine structures~\cite{Lin_2023_ICCV}; and InfoCD introduces an information-theoretic margin for discrimination~\cite{lin2023infocd}. 
While effective in some regimes, these methods do not explicitly encode \emph{coverage} (recall of surface regions) or penalize \emph{spurious predictions} (precision), leaving the recall--precision trade-off implicit. 
This gap motivates QAL, which directly incorporates coverage-aware weighting with an uncovered-region attraction signal.

\noindent\textbf{Thresholded coverage and precision metrics.}
Set coverage and F-score@$\tau$ are widely used to assess completeness and geometric accuracy of reconstructions~\cite{achlioptas2018learning,Tatarchenko_2019_CVPR}. 
We adopt thresholded Coverage (Cov@$\,\tau$) as a recall proxy and Spurious Points (SP@$\,\tau$) as a precision proxy, aligned with these practices and efficiently implemented on GPU. 
Unlike raw CD/EMD scores, these thresholded measures explicitly separate recall from precision, making them more interpretable for downstream evaluation. 
Our work closes the loop by introducing a loss function that is directly aligned with these thresholded metrics, rather than using them only for evaluation.

\noindent\textbf{Point-cloud and single-view 3D learning.}
Permutation-invariant encoders such as PointNet/PointNet++~\cite{qi_pointnet_2017,qiPointNetDeepHierarchical2017} introduced set-wise MLP+pool and hierarchical aggregation, while sparse 3D convolutions (Minkowski) enable large-scale voxelized scene processing~\cite{choyminkowski}. 
Decoders for shape completion and reconstruction typically follow a coarse-to-fine strategy: PCN expands seed points via folding~\cite{yuan_pcn_2018}, TopNet upsamples with a rooted tree~\cite{tchapmi_topnet_2019}, ECG refines predictions through error-correcting graphs~\cite{pan_ecg_2020}, and SeedFormer leverages attention for dense surface recovery~\cite{zhou2022seedformer}. 
Single-view pipelines extend this paradigm to images, ranging from early voxel-based models such as 3D-R2N2~\cite{choy20163d,choy_3d-r2n2_2016} to point-set generators~\cite{fan2017point} and more recent designs using transformers (LegoFormer) or multi-scale fusion (Pix2Vox/++)~\cite{yagubbayli2021legoformer,xie2019pix2vox,xie2020pix2vox++}. 
Although these approaches vary in representation and architecture, they almost universally optimize with CD or EMD for supervision, leaving them prone to holes and spurious clusters. 
This shared reliance makes them natural candidates for QAL, which directly encodes recall and precision while remaining compatible with existing pipelines.

\noindent\textbf{Robotics-driven shape completion.}
Robotics applications place stricter demands on recall and precision than generic 3D vision tasks. 
For manipulation and navigation, completing occluded geometry is critical: missing surfaces reduce grasp recall, while spurious points induce false positives. 
Systems either complete shapes before grasping~\cite{varley2017shape} or plan under uncertainty~\cite{lundell2020beyond}, and are typically evaluated with thresholded surface criteria such as F-score@$\tau$. 
Recent efforts such as GraspNet-1Billion~\cite{fang2020graspnet} emphasize the link between object geometry and grasp success, while learning-based scoring methods~\cite{morrison2020learning, liang2019pointnetgpd} show that fine-grained geometry directly impacts grasp reliability. 
These findings motivate our evaluation of QAL not only with surface metrics (Cov@$\,\tau$/SP@$\,\tau$) but also through grasp utility under pre-trained models.

\section{Quality Aware Loss Design}
\label{sec:ourwork}
In this section, we describe our proposed loss function \ourLoss{} (Quality-Aware Loss, QAL) and its role in 3D learning. QAL is broadly applicable across volumetric learning tasks and network structures. It captures differences between candidate and reference point clouds along multiple dimensions, enabling faster and more robust learning.

\subsection{Motivation} 
In most volumetric learning applications, Chamfer distance and its derivatives or Earth Movers Distance are the typical loss metrics used. But they optimize only average nearest-neighbor distances and therefore agnostic to \emph{holes} (low recall) and \emph{non-uniform densities} (low precision), yielding missing parts and spurious clusters. \cref{fig:intro} shows that CD-trained completions omit finer details in the chair despite low mean distance; Hyperbolic CD (HCD) and Info CD (InfoCD) reduce but do not eliminate such omissions (\cref{fig:psg_visualization}). %However, downstream reconstruction (e.g., meshing) might require uniform, well-covered point sets.
%Moreover, CD and its variants are \emph{blind to missing correspondences}: if a ground-truth region has no nearby predictions, the loss does not penalize it, leaving uncovered areas unaddressed.
The primary cause for this is that CD and its variants are \emph{blind to missing correspondences}: if a ground-truth region has no nearby predictions, the loss does not penalize it, leaving uncovered areas unaddressed.
However, several applications such as object recognition, robot localization, 3D tracking etc. require uniform coverage across the scene or object of interest. 
%With CD, predictions are often spread unevenly and cause spurious geometry (\cref{fig:psg_visualization}). 

\begin{figure}[t]
\centering
\begin{tikzpicture}[
  % gt/.style={draw=blue,fill=blue,shape=square*,minimum size=3.5pt},
  gt/.style={draw=blue,fill=blue,shape=rectangle,minimum size=3.5pt,inner sep=0pt},
  pred/.style={draw=green!50!black,fill=green!60!black,circle,minimum size=3.5pt},
  pred_art/.style={draw=gray!70,fill=gray!70,circle,minimum size=3.5pt},
  covfill/.style={fill=green!35,opacity=0.25},
  eps/.style={draw=black!50,dashed,opacity=0.8},
  % resarrow/.style={draw=red!80!black,very thick,-{Latex[length=2.5mm]}},
  % covarrow/.style={draw=orange!85!black,very thick,-{Latex[length=2.5mm]}},
  % attrarrow/.style={draw=purple!80!black,very thick,densely dashed,-{Latex[length=2.5mm]}}
  covarrow/.style={draw=orange!85!black,very thick},
  attrarrow/.style={draw=purple!80!black,very thick,densely dashed}
]

% ---------- Panel A: Coverage ----------
\begin{scope}[xshift=-2.9cm, local bounding box=panelA]
  % GT points
  \node[gt] (g1) at (0,0) {};
  \node[gt] (g2) at (1.8,0.6) {};
  \node[gt] (g3) at (1.0,1.6) {};
  % epsilon balls
  \draw[eps] (g1) circle (0.7);
  \draw[eps] (g2) circle (0.7);
  \draw[eps] (g3) circle (0.7);
  % Predictions
  \node[pred] (p1) at (0.5,0.2) {};
  \node[pred] (p2) at (1.1,0.5) {};
  \node[pred] (p3) at (0.8,1.2) {};
  \node[pred] (p4) at (2.5,1.7) {}; % outside coverage
  % Coverage arrows (GT -> NN pred within eps)
  \draw[covarrow] (g1) -- (p1);
  \draw[covarrow] (g2) -- (p2);
  \draw[covarrow] (g3) -- (p3);
% inside the Coverage scope
    \node[gt] (g4) at (2.4,-0.2) {};        % uncovered GT
    \draw[eps] (g4) circle (0.7);           % its epsilon-ball
    \draw[attrarrow] (g4) -- (p4);          % attraction towards nearest candidate (outside ε)
  
  % Title
  % \node[anchor=south west,font=\footnotesize] at (-0.15,2.2) {(a) Coverage \& Attraction};
\end{scope}

% ---------- Panel B: Thresholded Coverage & Artifacts ----------
\begin{scope}[xshift=1.5cm, local bounding box=panelB]
  % GT points
  \node[gt] (g1) at (0,0) {};
  \node[gt] (g2) at (1.8,0.6) {};
  \node[gt] (g3) at (1.0,1.6) {};
  \node[gt] (g4) at (2.1,-0.2) {}; % uncovered GT

  % Predictions (candidates)
  \node[pred]     (p1) at (0.45,0.15) {};
  \node[pred]     (p2) at (1.10,0.55) {};
  \node[pred]     (p3) at (0.85,1.25) {};
  \node[pred_art] (p4) at (2.5,1.5) {}; % artifact (outside all eps-balls)
  \node[pred_art] (p5) at (-0.3,1.0) {}; % another artifact

  % epsilon-balls (draw first, then fill the covered ones)
  \draw[eps] (g1) circle (0.70);
  \draw[eps] (g2) circle (0.70);
  \draw[eps] (g3) circle (0.70);
  \draw[eps] (g4) circle (0.70); % no fill: uncovered

  % fill green areas for GTs that are covered (have >=1 candidate within ε)
  \begin{scope}
    \fill[covfill] (g1) circle (0.70);
    \fill[covfill] (g2) circle (0.70);
    \fill[covfill] (g3) circle (0.70);
  \end{scope}

  % Optional arrows to indicate matches contributing to coverage
  \draw[covarrow] (g1) -- (p1);
  \draw[covarrow] (g2) -- (p2);
  \draw[covarrow] (g3) -- (p3);

  % Title
  % \node[anchor=south west,font=\footnotesize] at (-1.25,2.2) {(b) $\varepsilon$ - Coverage  \& Artifacts};
\end{scope}

\begin{scope}[on background layer]
  \draw[rounded corners=3pt,line width=0.6pt,black!40]
    ([xshift=-4pt,yshift=-4pt]panelA.south west)
    rectangle
    ([xshift=4pt,yshift=4pt]panelA.north east);
  \draw[rounded corners=3pt,line width=0.6pt,black!40]
    ([xshift=-4pt,yshift=-4pt]panelB.south west)
    rectangle
    ([xshift=4pt,yshift=4pt]panelB.north east);
\end{scope}

\node[anchor=south west,font=\footnotesize,fill=white,inner sep=1pt]
  at ([yshift=1pt]panelA.north west) {(a) Coverage \& Attraction};
\node[anchor=south west,font=\footnotesize,fill=white,inner sep=1pt]
  at ([yshift=1pt]panelB.north west) {(b) $\varepsilon$-Coverage \& Spurious Pts.};

% ---------- Legend (two rows, with attraction) ----------
\begin{scope}[yshift=-1.4cm, xshift=-2cm]
  % Row 1: markers
  \node[gt,label=right:{\footnotesize Reference (GT)}] at (-1.4,0) {};
  \node[pred,label=right:{\footnotesize Candidate}]     at (1.0,0)   {};
  \draw[eps] (3.0,0) circle (0.18) node[right=7pt]{\footnotesize $\varepsilon$-ball};
  \node[pred_art,label=right:{\footnotesize Spurious Pts.}]        at (4.5,0)   {};

  % Row 2: arrows
  \draw[covarrow]  (-1.7,-0.65) -- ++(0.7,0) node[right]{\footnotesize Coverage};
  % \draw[resarrow]  (1.2,-0.65)  -- ++(0.7,0) node[right]{\footnotesize Resolution};
  \draw[attrarrow] (1.2,-0.65)  -- ++(0.7,0) node[right]{\footnotesize Attraction (non-coverage)};
\end{scope}

\end{tikzpicture}
\vspace{-3mm}
\caption{\textbf{Illustration of QAL and evaluation metrics.} 
(a) QAL components: GT points (blue squares) with $\epsilon$-balls; coverage links (orange) connect GT$\rightarrow$prediction matches, while attraction links (purple) pull predictions toward uncovered GT regions. 
(b) Evaluation metrics: predictions inside $\epsilon$-balls are counted toward thresholded coverage, while those outside (gray) are spurious points.}
\vspace{-6mm}
\label{fig:illustrate}
\end{figure}
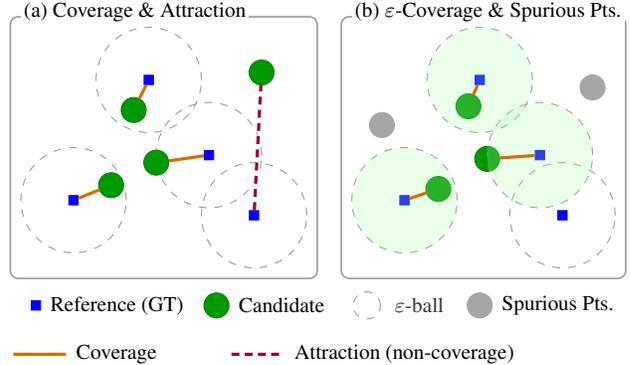

QAL addresses these limitations by combining two complementary ideas. First, it emphasizes regions that are under-covered, ensuring that thin or missing structures are recovered. Second, it explicitly attracts predictions toward ground-truth points that would otherwise remain unmatched, closing holes while preserving already accurate regions. The result is a more uniform and complete point set, aligning naturally with thresholded metrics such as Coverage and F1.
\cref{fig:illustrate}(a) provides intuition for our proposed Quality-Aware Loss (QAL). The coverage-weighted term \eqref{eq:l_cov} reweights nearest-neighbor distances with a soft margin $\epsilon$, down-weighting already well-matched predictions and emphasizing regions that remain under-covered. In parallel, the uncovered--GT attraction term \eqref{eq:l_attr} explicitly pulls predictions toward ground-truth points without nearby matches, directly eliminating holes and restoring missing structures. 

Together, these two components (orange coverage links and purple attraction links in \cref{fig:illustrate}(a)) encourage recall without disturbing already precise regions, producing point sets that are both more uniform and more complete. This design aligns optimization with commonly used metrics such as Coverage and F1, where missing geometry is typically more harmful than modest over-prediction. 

This intuition leads directly to QAL’s formulation, which combines coverage-weighted distances with uncovered--GT attraction into a tunable objective that explicitly balances recall and precision.

% QAL addresses these issues by (i) reweighting distances with a soft margin to emphasize under-covered regions, and (ii) attracting predictions toward uncovered GT points to explicitly drive recall. This design closes holes without disturbing already precise regions, aligning training with coverage/F-score metrics used in robotic shape completion and manipulation, where missing geometry is typically more harmful than modest over-prediction.
% \kar{Expand this a little more - describe the coverage/attraction using Figure 2 and the effect of such reasoning. This should lead into the exact definition of QAL in next subsection. Merge this subsection with 3.3}
% \noindent \cref{fig:illustrate} sketches the mechanism. Coverage-weighted distances (\eqref{eq:l_cov}) reduce average error where it matters most, while uncovered--GT attraction (\eqref{eq:l_attr}) eliminates missing structures. Together, they form a recall-oriented objective aligned with F1-style evaluation common in robotics tasks.

% \subsection{Intuition behind \ourLoss{}}

\subsection{Quality-Aware Loss (QAL)}
Let $A{=}\{a_i\}$ be predictions and $B{=}\{b_j\}$ ground truth. Denote nearest-neighbor distances by
$d_a = d(a,B)$ and $d_b = d(b,A)$ for brevity.

\noindent\textbf{Coverage-weighted term.}
We introduce a soft margin that emphasizes errors beyond a tolerance $\epsilon{>}0$:
\begin{equation}
w_{\mathrm{cov}}(d)=1-\big(\sigma(\omega(\epsilon-d))-0.5\big),\quad \omega{>}0,
\label{eq:wcov}
\end{equation}
where $\sigma(\cdot)$ is the logistic sigmoid. The weighted Chamfer-style loss is:
\begin{align}
\mathcal{L}_{\mathrm{cov}} &=
\frac{1}{|A|}\sum_{a\in A} w_{\mathrm{cov}}(d_a)\, d_a 
+ \frac{1}{|B|}\sum_{b\in B} w_{\mathrm{cov}}(d_b)\, d_b.
\label{eq:l_cov}
\end{align}
This term down-weights already precise matches and emphasizes errors outside the $\epsilon$-ball, aligning training with thresholded evaluation metrics such as F-score@1\%.

\noindent\textbf{Uncovered--GT attraction.}
Define the mask
\begin{equation}
m_B(b)=\mathbbm{1}\!\big[\,\#\{a\in A:\mathrm{NN}_B(a)=b\}=0\,\big],
\label{eq:mask}
\end{equation}
which marks ground-truth points $b$ that have no predicted neighbor (i.e., uncovered). Predictions are then attracted toward these regions:
\begin{equation}
\mathcal{L}_{\mathrm{attr}}
= \frac{1}{|B|}\sum_{b\in B} m_B(b)\,\sigma\!\big(\omega(\epsilon - d_b)\big)\, d_b.
\label{eq:l_attr}
\end{equation}
\cref{fig:illustrate}(b) uses a one-way illustration for simplicity; the implemented attraction operates symmetrically on uncovered GT regions.

\noindent\textbf{Final objective.}
\begin{equation}
\boxed{\;\mathcal{L}_{\mathrm{QAL}}
= \mathcal{L}_{\mathrm{cov}}
+ \lambda_{\mathrm{attr}}\,\mathcal{L}_{\mathrm{attr}},\quad
\lambda_{\mathrm{attr}}\ge 0\;}
\label{eq:l_qal}
\end{equation} 
where $\epsilon$ is the tolerance, $\omega$ controls sharpness, and $\lambda_{\mathrm{attr}}$ balances coverage and attraction. The overall computation is summarized in \cref{alg:qal}. 
In practice, $\epsilon$ is set to match evaluation tolerance (e.g., $1\%$ of object scale), $\omega$ regulates the smoothness of weighting, and $\lambda_{\mathrm{attr}}$ is selected via validation ablations (see \cref{sec:experiments}).

Together, the two terms promote coverage of missing structures while suppressing excessive clustering.

\noindent\textbf{Computational cost.}
Let $N{=}|A|$ and $M{=}|B|$. Chamfer Distance (two 1-NN passes) runs in $O(NM)$ with brute-force GPU \texttt{cdist}, or $O((N{+}M)\log(N{+}M))$ using kd-trees. QAL shares the same asymptotic cost, adding only $O(N{+}M)$ for per-point weights and masks~\cite{yuan_pcn_2018}. Thus, QAL has CD-like efficiency.

% \begin{algorithm}[t]
% \caption{Quality-Aware Loss (QAL) Computation}
% \label{alg:qal}
% \begin{algorithmic}[1]
% \REQUIRE Predicted set $A=\{a_i\}$, ground-truth set $B=\{b_j\}$, tolerance $\epsilon$, sharpness $\omega$, weight $\lambda_{\mathrm{attr}}$
% \STATE Compute nearest-neighbor distances: $d_a = d(a,B)$ for $a \in A$, and $d_b = d(b,A)$ for $b \in B$
% \STATE Compute coverage weights: $w_{\mathrm{cov}}(d) = 1 - (\sigma(\omega(\epsilon-d)) - 0.5)$
% \STATE Coverage term: $\mathcal{L}_{\mathrm{cov}} = \tfrac{1}{|A|}\sum_{a} w_{\mathrm{cov}}(d_a) d_a + \tfrac{1}{|B|}\sum_{b} w_{\mathrm{cov}}(d_b) d_b$
% \STATE Mark uncovered GT points: $m_B(b) = 1$ if no $a \in A$ has $\mathrm{NN}_B(a)=b$, else $0$
% \STATE Attraction term: $\mathcal{L}_{\mathrm{attr}} = \tfrac{1}{|B|}\sum_{b} m_B(b)\, \sigma(\omega(\epsilon-d_b)) d_b$
% \STATE Final objective: $\mathcal{L}_{\mathrm{QAL}} = \mathcal{L}_{\mathrm{cov}} + \lambda_{\mathrm{attr}} \mathcal{L}_{\mathrm{attr}}$
% \RETURN $\mathcal{L}_{\mathrm{QAL}}$
% \end{algorithmic}
% \end{algorithm}

\begin{algorithm}[t]
\caption{Quality-Aware Loss (QAL)}
\label{alg:qal}
\begin{algorithmic}[1]
\REQUIRE Predicted set $A$, ground truth $B$, tolerance $\epsilon$, sharpness $\omega$, weight $\lambda_{\mathrm{attr}}$
\STATE Compute NN distances: $d_a=d(a,B)$ for $a\!\in\!A$, $d_b=d(b,A)$ for $b\!\in\!B$
\STATE Coverage: $\mathcal{L}_{\mathrm{cov}}=\tfrac{1}{|A|}\sum_{a}w(d_a)d_a+\tfrac{1}{|B|}\sum_{b}w(d_b)d_b$, where $w(d)=1-(\sigma(\omega(\epsilon-d))-0.5)$
\STATE Attraction: $\mathcal{L}_{\mathrm{attr}}=\tfrac{1}{|B|}\sum_{b} m_B(b)\,\sigma(\omega(\epsilon-d_b))d_b$, with $m_B(b){=}1$ if $b$ unmatched
\STATE Final: $\mathcal{L}_{\mathrm{QAL}}=\mathcal{L}_{\mathrm{cov}}+\lambda_{\mathrm{attr}}\mathcal{L}_{\mathrm{attr}}$
\RETURN $\mathcal{L}_{\mathrm{QAL}}$
\end{algorithmic}
\end{algorithm}

% \vspace{-10pt}
\section{Experiments}
\label{sec:experiments}

We evaluate QAL across three complementary settings spanning controlled loss comparisons, large-scale retraining, and task-driven evaluation. 
First, a micro-benchmark on MVP~\cite{pan2021variational} compares QAL against CD, EMD, and recent variants under identical backbones, isolating the effect of the loss itself. 
Second, we perform full retraining on PCN~\cite{yuan_pcn_2018} and ShapeNet-55~\cite{lin2023infocd} to test generalization across datasets and architectures at higher output resolution. 
Third, we assess downstream task utility using GraspNet~\cite{fang2020graspnet}, where completions trained with QAL are evaluated via grasp quality scores. 

For evaluation, we report Chamfer Distance (CD) and F1 as standard metrics, and additionally introduce \emph{Coverage} and \emph{Spurious Points} to provide a multidimensional assessment of recall–precision balance. We also conduct ablations on hyperparameters and output resolution to study stability.

\noindent\textbf{Datasets and Test Pipelines.} 
We evaluate across three representative tasks: point cloud completion, image-to-3D reconstruction, and image-to-mesh reconstruction. 
\textbf{MVP}~\cite{pan2021variational} provides multi-view partial point clouds of 16 categories with 62k training and 41k testing pairs. Partial shapes are rendered from ShapeNet CAD models, and ground-truth point clouds are sampled via Poisson Disk Sampling. 
The \textbf{PCN} benchmark~\cite{yuan_pcn_2018} covers 8 ShapeNet categories, where partial inputs are obtained from depth back-projections; we follow the standard train/test split. 
\textbf{ShapeNet-55}~\cite{lin2023infocd} uses hole/occlusion synthesis under three difficulty levels (S/M/H); we report L2-CD$\times 1e^{3}$ and F1@\,$1\%$ following prior work.

\noindent\textbf{Training Setup.} 
For the \textbf{MVP} micro-benchmark, all networks were trained for 100 epochs with batch size 32 and learning rate \(1\times10^{-3}\).
We set \(\omega=10.0\) from ablations, \(\epsilon=0.001\) (1\,cm for meter-scale objects), and \(\lambda_{\mathrm{attr}}=1.0\). 
Coverage is reported at $\epsilon=0.03$, approximately twice the average nearest-neighbor spacing at 2048 points (see Supplementary for details). 
Nearest-neighbor distance is Euclidean (\(\ell_2\)) in 3D, i.e., \(d(x,Y)=\min_{y\in Y}\|x-y\|_2\). 
For fairness, we use official implementations and training protocols, changing only the loss. 
To highlight coverage under limited capacity, we set the output resolution to \textbf{2048 points} for MVP. 
For \textbf{PCN} and \textbf{ShapeNet-55}, we perform \emph{full retraining} with the original network settings, including \textbf{16K output points}, learning rate, batch size, and balance factors. 
All experiments were run on a workstation with an Intel i9-12900K, 64\,GB RAM, and 2 NVIDIA RTX A6000 GPUs.

\noindent\textbf{Evaluation Metrics.}
In addition to CD, EMD, and F1~\cite{Tatarchenko_2019_CVPR}, we introduce shape-level, thresholded metrics to better expose recall–precision trade-offs. 
Let $d(b,A)$ be the nearest-neighbor distance from $b\!\in\!B$ (GT) to $A$, and $d(a,B)$ analogously. We define:
\begin{equation}
\begin{aligned}
\mathrm{Cov}@\tau &= \tfrac{1}{|B|}\sum_{b\in B}\mathbf{1}[d(b,A)\le\tau],\\
\mathrm{SP}@\tau  &= \tfrac{1}{|A|}\sum_{a\in A}\mathbf{1}[d(a,B)>\tau].
\end{aligned}
\label{eq:cov_sp}
\end{equation}
% Here, $\mathrm{Cov}$ measures recall of GT surface and $\mathrm{SP}$ measures spurious predictions. 
% For comparability, we use $\overline{\mathrm{SP}}\!=1-\mathrm{SP}$ (higher-is-better), and report percentages for both. 
% We also define an aggregate quality score as $\mathrm{Quality}=\tfrac{1}{2}(\mathrm{Cov}+\overline{\mathrm{SP}})$, though in the main text we emphasize per-metric reporting (see Supplementary for grouped-bar visualizations).

Here, $\mathrm{Cov}$ measures recall of the GT surface and $\mathrm{SP}$ measures spurious predictions; \cref{fig:illustrate}(b) provides a visual illustration of the $\epsilon$-ball interpretation used for both metrics. 
For comparability, we use $\overline{\mathrm{SP}}\!=1-\mathrm{SP}$ (higher-is-better), and report percentages for both. 
We also define an aggregate quality score as $\mathrm{Quality}=\tfrac{1}{2}(\mathrm{Cov}+\overline{\mathrm{SP}})$, though in the main text we emphasize per-metric reporting (see Supplementary for grouped-bar visualizations).

% \noindent\textbf{Reporting trade-offs.}
% Recall-oriented training naturally induces a trade-off between coverage and precision. 
% To make this explicit, our ablations present \emph{grouped bar plots} showing side-by-side 
% \(\Cov\) ($\uparrow$), the complement of spurious points $\overline{\mathrm{SP}}$ ($\uparrow$), and CD ($\downarrow$) 
% for each hyperparameter setting (\cref{fig:qal_hyperparameter_ablation}). 
% We avoid collapsing these metrics into a single score in the main text; instead we emphasize 
% per-metric reporting to surface the trade-offs directly.

\begin{figure*}[!th]
  \centering
  \vspace{-1cm}
\includegraphics[width=0.8\textwidth]{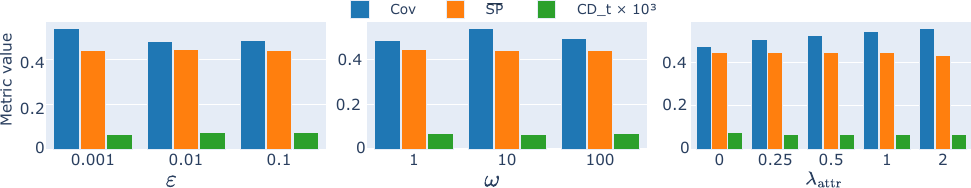}
\vspace{-0.2cm}
  \caption{\textbf{QAL hyperparameter ablation.} Each panel shows grouped bars for Coverage (Cov; $\uparrow$), Spurious Points (\(\overline{\mathrm{SP}}\); $\uparrow$), and Chamfer Distance (CD$\times10^{3}$; $\downarrow$). 
\emph{Left:} $\epsilon$ sweep with $\omega\!\approx\!10$, $\lambda_{\mathrm{attr}}\!\approx\!1.0$. 
\emph{Middle:} $\omega$ sweep at the selected $\epsilon^{\star}$. 
\emph{Right:} $\lambda_{\mathrm{attr}}$ sweep at $(\epsilon^{\star},\omega^{\star})$. }
% Stars $(\epsilon^{\star},\omega^{\star})$ are chosen by maximizing Cov and then minimizing SP (ties broken by CD). 
% Shared legend on top; common $y$-axis at left.}
  \label{fig:qal_hyperparameter_ablation}
  \vspace{-0.4cm}
\end{figure*}
\subsection{Ablation Studies}

% We ablate QAL incrementally in three stages (~\cref{fig:qal_hyperparameter_ablation}): 

% \noindent(1) \textbf{$\epsilon$ sweep} with $\omega{\approx}10$, $\lambda_{\mathrm{attr}}{\approx}0.0$ to select $\epsilon^\star$ by maximizing Cov@$\epsilon$ (ties by minimizing SP@$\epsilon$, then CD); (2) \textbf{$\omega$ sweep} at $\epsilon^\star$ with the same criterion to get $\omega^\star$; (3) \textbf{$\lambda_{\mathrm{attr}}$ sweep} at $(\epsilon^\star,\omega^\star)$ over $\{0,0.25,0.5,1.0,2.0\}$. Plots show grouped bars (Cov$\uparrow$/$\overline{\mathrm{SP}}$$\uparrow$/CD$\downarrow$) for each setting; CD is scaled by $10^3$ for readability.

We ablate QAL incrementally in three stages (~\cref{fig:qal_hyperparameter_ablation}): 
(1) \textbf{$\epsilon$ sweep} with $\omega{\approx}10$, $\lambda_{\mathrm{attr}}{\approx}0.0$ to select $\epsilon^\star$ by maximizing Cov@$\epsilon$ (ties by minimizing SP@$\epsilon$, then CD;\ $\epsilon<10^{-3}$ gave unstable gradients and diminishing returns), 
(2) \textbf{$\omega$ sweep} at $\epsilon^\star$ with the same criterion to obtain $\omega^\star$, and 
(3) \textbf{$\lambda_{\mathrm{attr}}$ sweep} at $(\epsilon^\star,\omega^\star)$ over $\{0,0.25,0.5,1.0,2.0\}$. 
Plots show grouped bars (Cov$\uparrow$/$\overline{\mathrm{SP}}$$\uparrow$/CD$\downarrow$), with CD scaled by $10^3$ for readability.

\noindent\textbf{Observations.}
From \cref{fig:qal_hyperparameter_ablation}, $\epsilon{=}0.001$ yields the best trade-off (high Cov, high $\cSP$), while $\epsilon{=}0.1$ and $\epsilon{=}0.01$ underfit the regions; $\epsilon<10^{-3}$ further leads to unstable gradients and diminishing returns. At $\epsilon^\star$, a moderate sharpness $\omega{=}10$ consistently outperforms $\omega{=}1$ (weak margin) and $\omega{=}100$ (overly sharp, slight \emph{decrease} in $\cSP$). Finally, at $(\epsilon^\star,\omega^\star)$, increasing $\lambda_{\mathrm{attr}}$ raises Cov but reduces $\cSP$; accordingly, $\mathrm{CD}\!\times\!10^3$ \emph{anti-correlates} with $\cSP$ (lower $\cSP$ $\Rightarrow$ higher CD) since spurious predictions inflate nearest-neighbor distances. We therefore pick the Pareto knee at $\lambda_{\mathrm{attr}}\!\in\![0.5,1.0]$, capturing most of the Cov gain while limiting the erosion of $\cSP$ and controlling CD. The exact point can be tuned to application tolerance for false positives. A small Optuna-based search on the lamp category further confirmed that the three hyperparameters are only weakly coupled, exhibiting a broad plateau around $(\epsilon{=}10^{-3},\omega{=}10,\lambda_{\mathrm{attr}}\!\in[0.5,1.0])$.

\subsection{Point cloud completion}
 % \vspace{-1cm}
\label{subsec:compnet}

\begin{table}[t]
%\vspace{-2.5em}
\centering
\caption{Completion Networks trained with CD (L2-CD $\times 1e^{3}$), HCD, InfoCD and \ourLoss{} for \(\epsilon =0.03\). \Cov{}, $\cSP$ and  Quality metrics are scaled as percentages
}
% \pranay{Update these results}
\vspace{-0.2cm}

\centering
\scriptsize
\setlength{\tabcolsep}{6pt}
\label{tab:qal_pcn_evaluation}
\begin{tabular}{l|cc|ccc}
\toprule
Method(Network+Loss) & CD \(\downarrow\) & \(F1 \uparrow\) & Cov. \(\uparrow\) & \(\overline{SP}\) \(\uparrow\) & Quality \(\uparrow\) \\
\midrule
PCN \cite{yuan_pcn_2018} + CD & 17.4 & 0.29 & 59.8 & 64.0 & 61.9 \\
PCN + EMD & 45.2 & 0.11 & 21.1 & 40.9 & 31.0 \\
PCN + DCD & 17.5 & 0.29 & 63.2 & 62.9 & 63.1 \\
PCN + HCD & 17.5 & \textbf{0.30} & 59.8 & \textbf{64.3} & 62.1 \\
PCN + INFOCD & \textbf{17.2} & \textbf{0.30} & 62.8 & 62.5 & 62.6 \\
PCN + QAL & \textbf{17.2} & 0.29 & \textbf{65.9} & 61.4 & \textbf{63.6} \\
\cmidrule{1-6}
TOP-NET \cite{tchapmi_topnet_2019} + CD & \textbf{18.1} & \textbf{0.27} & 55.2 & \textbf{61.0} & 58.1 \\
TOP-NET + HCD & 18.3 & \textbf{0.27} & 55.8 & 60.0 & 57.9 \\
TOP-NET + INFOCD & 18.2 & \textbf{0.27} & 56.3 & 60.3 & 58.3 \\
TOP-NET + QAL & 18.3 & 0.26 & \textbf{59.1} & 57.7 & \textbf{58.4} \\
\cmidrule{1-6}
ECG~\cite{pan_ecg_2020} + CD & 14.0 & \textbf{0.45} & 66.6 & 72.1 & 69.3 \\
ECG + EMD & 15.5 & 0.38 & 67.0 & 70.1 & 68.5 \\
ECG + DCD & 14.0 & \textbf{0.45} & 68.1 & 71.6 & 69.8 \\
ECG + HCD & 14.0 & \textbf{0.45} & 66.8 & \textbf{72.4} & 69.6 \\
ECG + INFOCD & 14.0 & \textbf{0.45} & 67.0 & 71.2 & 69.1 \\
ECG + QAL & \textbf{13.8} & \textbf{0.45} & \textbf{68.9} & 71.3 & \textbf{70.0} \\
\cmidrule{1-6}
Seedformer \cite{zhou2022seedformer} + CD & \textbf{12.3} & \textbf{0.51} & 73.0 & 80.8 & \textbf{76.9} \\
Seedformer + INFOCD & 14.7 & 0.41 & 53.4 & \textbf{85.5} & 69.5 \\
Seedformer + QAL & 12.9 & 0.47 & \textbf{77.8} & 74.4 & 76.1 \\
\bottomrule
\end{tabular}

\vspace{-0.4cm}
% \vspace{-0.5cm}
\end{table}
% Table 1: PCN (per-point L1-CD ×1000)
\begin{table}[t]\centering\small
\caption{Comparison on PCN in terms of L1-CD $\times 1e^{3}$.}
\vspace{-0.2cm}
\centering
\scriptsize
\setlength{\tabcolsep}{1pt}

\begin{tabular}{c|rrrrrrrr|r}
\toprule
Methods & Plane & Cabinet & Car & Chair & Lamp & Couch & Table & Boat & Avg. \\
\midrule
TopNet~\cite{tchapmi_topnet_2019} & 7.61 & 13.31 & 10.90 & 13.82 & 14.44 & 14.78 & 11.22 & 11.12 & 12.15 \\
% AtlasNet~\cite{atlasnet} & 6.37 & 11.94 & 10.10 & 12.06 & 12.37 & 12.99 & 10.33 & 10.61 & 10.85 \\
% GRNet~\cite{grnet} & 6.45 & 10.37 & 9.45 & 9.41 & 7.96 & 10.51 & 8.44 & 8.04 & 8.83 \\
% CRN~\cite{crn} & 4.79 & 9.97 & 8.31 & 9.49 & 8.94 & 10.69 & 7.81 & 8.05 & 8.51 \\
% NSFA~\cite{nsfa} & 4.76 & 10.18 & 8.63 & 8.53 & 7.03 & 10.53 & 7.35 & 7.48 & 8.06 \\
% FBNet~\cite{fbnet} & 3.99 & 9.05 & 7.90 & 7.38 & 5.82 & 8.85 & 6.35 & 6.18 & 6.94 \\
\cmidrule{1-10}
PCN~\cite{yuan_pcn_2018} & 5.50 & 22.70 & 10.63 & 8.70 & 11.00 & 11.34 & 11.68 & 8.59 & 11.27 \\
% HyperCD{+}PCN & 5.95 & 11.62 & 9.33 & 12.45 & 12.58 & 13.10 & 9.82 & 9.85 & 10.59 \\
InfoCD{+}PCN & \textbf{5.07} & 22.27 & 10.18 & \textbf{8.26} & \textbf{10.57} & \textbf{10.98} & 11.23 & \textbf{8.15} & 10.83 \\
QAL{+}PCN & 5.92 & \textbf{11.09} & \textbf{9.12} & 11.99 & 12.70 & 12.36 & \textbf{9.70} & 10.42 & \textbf{10.41} \\
\cmidrule{1-10}
FoldingNet~\cite{yang2018foldingnet} & 9.49 & 15.80 & 12.61 & 15.55 & 16.41 & 15.97 & 13.65 & 14.99 & 14.31 \\
% HyperCD{+}FoldingNet & 7.89 & 12.90 & 10.67 & 14.55 & 13.87 & 14.09 & 11.86 & 10.89 & 12.09 \\
InfoCD{+}FoldingNet & 7.90 & \textbf{12.68} & \textbf{10.83} & 14.04 & 14.05 & 14.56 & \textbf{11.61} & 11.45 & 12.14 \\
QAL{+}FoldingNet & \textbf{7.62} & 13.47 & 11.24 & \textbf{14.00} & \textbf{13.36} & \textbf{14.41} & 11.75 & \textbf{10.81} & \textbf{12.08} \\
\cmidrule{1-10}
PMP\mbox{-}Net~\cite{wen2021pmp} & 5.65 & 11.24 & 9.64 & 9.51 & 6.95 & 10.83 & 8.72 & 7.25 & 8.73 \\
% HyperCD{+}PMP\mbox{-}Net & 5.06 & 10.67 & 9.30 & 9.11 & 6.83 & 11.01 & 8.18 & 7.03 & 8.40 \\
InfoCD{+}PMP\mbox{-}Net & 4.67 & 10.09 & 8.87 & 8.59 & 6.38 & 10.48 & 7.51 & 6.75 & 7.92 \\
QAL{+}PMP\mbox{-}Net & \textbf{4.51} & \textbf{9.93} & \textbf{8.69} & \textbf{8.29} & \textbf{6.20} & \textbf{9.86} & \textbf{7.31} & \textbf{6.53} & \textbf{7.66} \\
\cmidrule{1-10}
% PoinTr~\cite{yu2021pointr} & 4.75 & 10.47 & 8.68 & 9.39 & 7.75 & 10.93 & 7.78 & 7.29 & 8.38 \\
% HyperCD{+}PoinTr & 4.42 & 9.77 & 8.22 & 8.22 & 6.62 & 9.62 & 6.97 & 6.67 & 7.56 \\
% InfoCD{+}PoinTr & 4.06 & 9.42 & 8.11 & 7.81 & 6.21 & 9.38 & 6.57 & 6.40 & 7.24 \\
% \cmidrule{1-10}
% SnowflakeNet~\cite{snowflakenet} & 4.29 & 9.16 & 8.08 & 7.89 & 6.07 & 9.23 & 6.55 & 6.40 & 7.21 \\
% HyperCD{+}SnowflakeNet & 3.95 & 9.01 & 7.88 & 7.37 & 5.75 & 8.94 & 6.19 & 6.17 & 6.91 \\
% InfoCD{+}SnowflakeNet & 4.01 & 8.81 & 7.62 & 5.51 & 5.80 & 8.91 & 6.21 & 5.05 & 6.86 \\
% PointAttN~\cite{pointattn} & 3.87 & 9.00 & 7.63 & 7.43 & 5.90 & 8.68 & 6.32 & 6.09 & 6.86 \\
% HyperCD{+}PointAttN & 3.76 & 8.93 & 7.49 & 7.06 & 5.61 & 8.48 & 6.25 & 5.92 & 6.68 \\
% InfoCD{+}PointAttN & 3.72 & 8.87 & 7.46 & 7.02 & 5.60 & 8.45 & 6.23 & 5.92 & 6.65 \\
SeedFormer~\cite{zhou2022seedformer} & 3.85 & 9.05 & 8.06 & 7.06 & 5.21 & 8.85 & 6.05 & 5.85 & 6.74 \\
% HyperCD{+}SeedFormer & 3.72 & 8.71 & 7.79 & 6.83 & 5.11 & 8.61 & 5.82 & 5.76 & 6.54 \\
% InfoCD{+}SeedFormer & \textbf{3.69} & \textbf{8.72} & \textbf{7.68} & \textbf{6.84} & \textbf{5.08} & \textbf{8.61} & \textbf{5.83} & \textbf{5.75} & \textbf{6.52} \\
QAL{+}SeedFormer & \textbf{3.83} & \textbf{8.89} & \textbf{7.80} & \textbf{6.99} & 5.26 & \textbf{8.72} & \textbf{5.91} & \textbf{5.84} & \textbf{6.65} \\
\bottomrule
\end{tabular}
\label{tab:pcn}

\vspace{-0.4cm}
\end{table}

% Table 3: ShapeNet-55 (L2-CD×1000 and F1)
\begin{table}[t]\centering\small
\caption{Results on ShapeNet\mbox{-}55 using L2\mbox{-}CD$\times 1e^{3}$ and F1 score. }
\vspace{-0.2cm}
\centering
\scriptsize
\setlength{\tabcolsep}{1pt}
\begin{tabular}{c|rrrrr|rrrr|r}
\toprule
Methods & Table & Chair & Plane & Car & Sofa & CD\mbox{-}S & CD\mbox{-}M & CD\mbox{-}H & Avg. & F1 \\
\midrule
% PFNet~\cite{pfnet} & 3.95 & 4.24 & 1.81 & 2.53 & 3.34 & 3.83 & 3.87 & 7.97 & 5.22 & 0.339 \\
% TopNet~\cite{tchapmi_topnet_2019} & 2.21 & 2.53 & 1.14 & 2.18 & 2.36 & 2.26 & 2.16 & 4.30 & 2.91 & 0.126 \\
\cmidrule{1-11}
PCN~\cite{yuan_pcn_2018} & 2.13 & 2.29 & 1.02 & 1.85 & 2.06 & 1.94 & \textbf{1.96} & 4.08 & 2.66 & 0.133 \\
QAL {+} PCN & \textbf{1.85} & \textbf{2.05} & \textbf{0.93} & \textbf{1.48} & \textbf{1.68} & \textbf{1.78} & 2.06 & \textbf{3.07} & \textbf{2.31} & \textbf{0.187} \\
\cmidrule{1-11}
% GRNet~\cite{grnet} & 1.63 & 1.88 & 1.02 & 1.64 & 1.72 & 1.35 & 1.71 & 2.85 & 1.97 & 0.238 \\
% FoldingNet~\cite{yang2018foldingnet} & 2.53 & 2.81 & 1.43 & 1.98 & 2.48 & 2.67 & 2.66 & 4.05 & 3.12 & 0.082 \\
% InfoCD{+}FoldingNet & 2.14 & 2.37 & 1.03 & 1.55 & 2.04 & 2.17 & 2.50 & 3.46 & 2.71 & 0.137 \\
% \cmidrule{1-11}
PoinTr~\cite{yu2021pointr} & 0.81 & 0.95 & 0.44 & 0.91 & 0.79 & 0.58 & 0.88 & 1.79 & 1.09 & 0.464 \\
% InfoCD{+}PoinTr & 0.69 & 0.83 & 0.33 & 0.80 & 0.67 & 0.47 & 0.73 & 1.50 & 0.90 & 0.524 \\
QAL{+}PoinTr & \textbf{0.74} & \textbf{0.87} & \textbf{0.42} & \textbf{0.82} & \textbf{0.72} & \textbf{0.49} & \textbf{0.79} & \textbf{1.64} & \textbf{0.98} & \textbf{0.52} \\
% \cmidrule{1-11}
% SeedFormer~\cite{zhou2022seedformer} & 0.72 & 0.81 & 0.40 & 0.89 & 0.71 & 0.50 & 0.77 & 1.49 & 0.92 & 0.472 \\
% HyperCD{+}SeedFormer & 0.66 & 0.74 & 0.35 & 0.83 & 0.64 & 0.47 & 0.72 & 1.40 & 0.86 & 0.482 \\
% InfoCD{+}SeedFormer & 0.65 & 0.72 & 0.31 & 0.81 & 0.62 & 0.43 & 0.71 & 1.38 & 0.84 & 0.490 \\
% QAL{+}SeedFormer & 0.76 & 0.88 & 0.42 & 0.91 & 0.73 & \textbf{0.47} & \textbf{0.70} & \textbf{1.30} & \textbf{0.82} & - \\
\bottomrule
\end{tabular}
\label{tab:shapenet55}
\vspace{-0.5cm}
\end{table}

\begin{figure*}[t]
  \centering
  % \vspace{-1cm}
  % \includegraphics[trim=2cm 1.2cm 2cm 1.2cm,clip=true,width=.9\textwidth]{images/QAL_quantitative_results.pdf}
  \includegraphics[width=.95\textwidth]{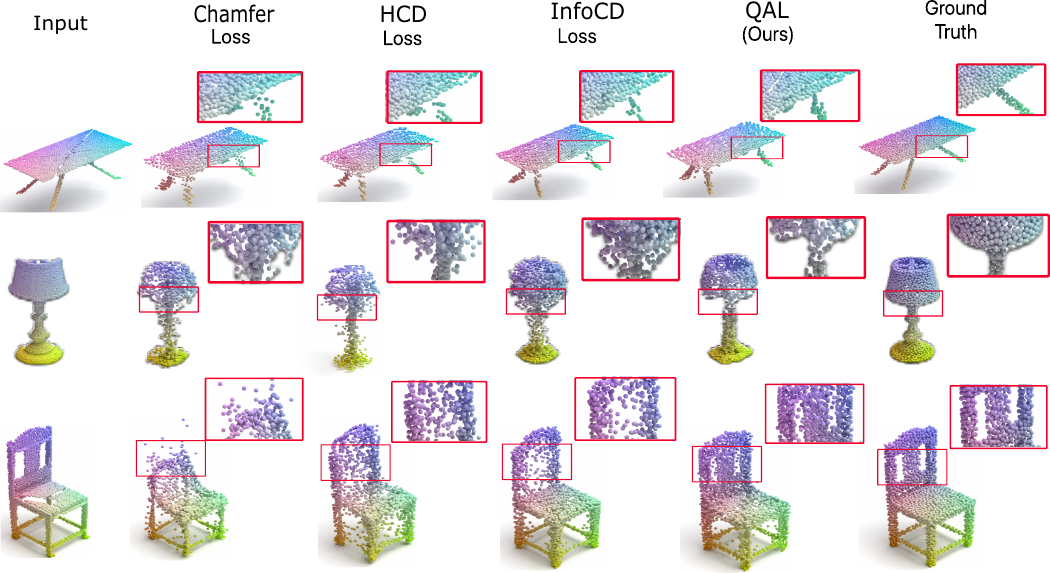}
  \vspace{-0.3cm}
  \caption{Qualitative results for point cloud completion for samples trained on Topnet, PCN, and ECG (Top to bottom) with four different loss functions CD, HCD, InfoCD, and \ourLoss{}. The highlighted column with \ourLoss{} learned more intricate details which are missed in models trained by other loss. These results are on the validation set after 100 epochs of training with  \(\epsilon=0.001\), \(\omega=10.0\) and \(\lambda_{attr}=1.0\)
}
  \label{fig:pcn_visualization}
  \vspace{-0.5cm}
\end{figure*}

PCN \cite{yuan_pcn_2018}, TopNet\cite{tchapmi_topnet_2019},  ECG\cite{pan_ecg_2020} and Seedformer\cite{zhou2022seedformer} were chosen for evaluation thanks to the availability of code and ease of modification to apply \ourLoss{}. All models were trained with the same random seed values to ensure invariance to run-to-run deviations. 
The only modification applied is the change in the loss function to \ourLoss{} which served as a drop-in replacement. Apart from the change in loss, the networks were trained as intended by their respective authors.

% \noindent\textbf{MVP dataset benchmarking (}\(\tau{=}0.03m\)\textbf{).}
% Coverage, a proxy for recall, increases across all backbones when trained with \ourLoss{} (\cref{tab:qal_pcn_evaluation}), with average gains of \textbf{+3.9 pts} (PCN: +6.1, TopNet: +3.9, ECG: +2.0, SeedFormer: +4.8). On PCN (+1.7), TopNet (+0.3), and ECG (+0.6) this also raises the composite \emph{Quality}, indicating that recall gains are achieved without loss of precision. On SeedFormer, Quality drops slightly (–0.8) despite strong Coverage improvements, reflecting the intended recall–precision trade. L2--CD shows small backbone-dependent shifts (lower for PCN/ECG, slightly higher for TopNet/SeedFormer), while F1 remains largely unchanged. Overall, \ourLoss{} consistently improves recall and often Quality, aligning training with completion behavior desired in practice. \textit{For completeness:} on \emph{TopNet}, EMD and DCD are omitted because both failed to converge across multiple seeds under author-recommended settings.
\noindent\textbf{MVP benchmarking and qualitative analysis (}\(\tau{=}0.03m\)\textbf{).} 
Coverage, a proxy for recall, increases across all backbones when trained with \ourLoss{} (\cref{tab:qal_pcn_evaluation}), with average gains of \textbf{+4.3 pts} (PCN: +6.1, TopNet: +3.9, ECG: +2.3, SeedFormer: +4.8). 
On PCN (+1.7), TopNet (+0.3), and ECG (+0.7) this also raises the composite \emph{Quality}, showing that recall gains are achieved with a moderate recall-oriented trade-off; %without loss of precision;
SeedFormer shows a slight Quality drop (–0.8), reflecting the intended recall–precision trade-off. 
L2--CD shifts are backbone-dependent (lower for PCN/ECG, slightly higher for TopNet/SeedFormer), while F1 remains largely unchanged. 
Overall, \ourLoss{} consistently improves recall and often Quality, aligning training with completion behavior desired in practice. 

\noindent\cref{fig:pcn_visualization} complements these results with qualitative comparisons for TopNet, PCN, and ECG under CD, HCD, InfoCD, and \ourLoss{}. 
SeedFormer is omitted here since we trained it only with CD, InfoCD, and \ourLoss{} due to its high computational cost, using it mainly for state-of-the-art comparison. 
Across the other backbones, \ourLoss{} produces higher coverage with sharper details than CD and HCD. 
CD often misses regions (e.g., ECG), HCD shows reduced recall (e.g., TopNet), and InfoCD, while competitive, still fails to recover fine structures. 
Crucially, the slight Coverage gains of \ourLoss{} are not only reflected quantitatively but also yield much more impactful qualitative improvements—recovering sharper edges and finer details for more complete reconstructions.
These results suggest that the benefits of \ourLoss{} extend beyond standard geometric metrics. 
Since sharper edges and improved coverage are especially critical for downstream tasks, we next evaluate how completions from these models influence robotic manipulation performance under a pre-trained GraspNet model (\cref{sec:grasp_utility}).

\noindent\textbf{PCN (L1-CD$\times 1e^{3}$).}
\ourLoss{} improves most where CD tends to \emph{accept holes}: bulky parts (e.g., \emph{Cabinet}) and thin appendages. The coverage-weighted term down-weights already-correct pairs inside the $\epsilon$ margin, so optimization budget shifts to regions with larger residuals; the uncovered–GT attraction then pulls points into missing areas, preventing the “average-to-the-middle” behavior typical of CD. Coarse-to-fine decoders (PCN, FoldingNet) benefit strongly because their early coarse sets leave sizeable uncovered regions that QAL explicitly targets. Seed-growing models (PMP-Net, SeedFormer) see steadier but smaller gains since they already densify surfaces; QAL still reduces residual and over-concentration by emphasizing uncovered GT. Compared to InfoCD, QAL’s uncovered-mask provides a more direct recall signal, which explains its higher gains on categories with large occlusions while remaining competitive elsewhere.

\noindent\textbf{ShapeNet\mbox{-}55 (L2-CD$\times 1e^{3}$ and F1).}
On this benchmark, the trends mirror PCN: CD decreases together with higher F1 when training with \ourLoss{}. This follows from QAL’s alignment with thresholded matching (F-score@$\epsilon$): emphasizing errors beyond $\epsilon$ improves the count of within-tolerance matches, which directly lifts F1. Improvements are strongest on the medium/hard splits, where large missing parts are to be completed; here, the attraction term closes gaps that CD leaves unresolved. Architectures that already capture fine detail (e.g., PoinTr) still gain because QAL deprioritizes easy pairs and suppresses spurious clusters just outside the tolerance, converting small geometric fixes into consistent F1 increases.

Across datasets and architectures, \ourLoss{} consistently lowers Chamfer Distance and raises F1, with the strongest effects on harder splits and occluded/thin structures—aligning with its coverage-first, SP-aware design.

% \textbf{Qualitative results:} 
% \cref{fig:pcn_visualization} shows qualitative comparisons for TopNet, PCN, and ECG under CD, HCD, InfoCD, and \ourLoss{}. 
% SeedFormer is omitted here since we trained it only with CD, InfoCD, and \ourLoss{} due to its high computational cost, using it primarily to compare against state-of-the-art rather than across all losses. 
% Across the other backbones, \ourLoss{} produces higher coverage with sharper details than CD and HCD. 
% CD often misses regions (e.g., ECG), while HCD exhibits reduced recall (e.g., TopNet). 
% InfoCD performs competitively but still fails to recover certain fine structures, whereas the slight coverage improvements of \ourLoss{} are not only reflected quantitatively but also yield much more impactful qualitative gains—recovering sharper edges and finer details for more complete reconstructions.

\begin{figure}[t]
     \centering
     \includegraphics[width=\linewidth]{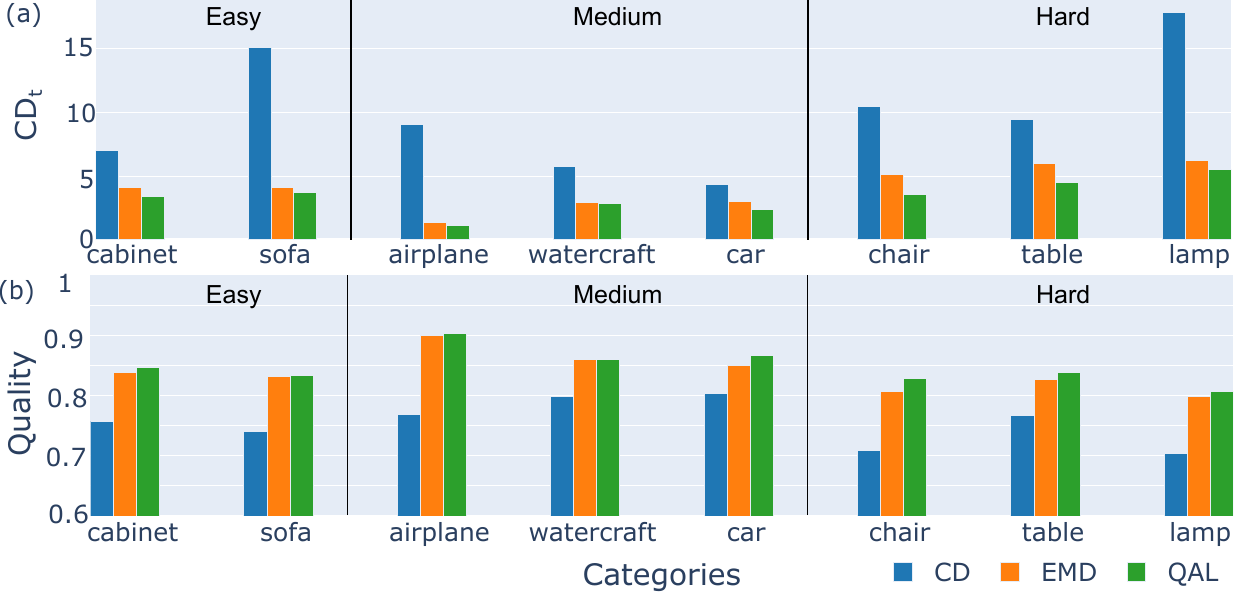}
\vspace{-0.5cm}
        \caption{Per-category metric scores comparison for ECG model trained with three loss functions: CD, EMD, and \ourLoss{}. Metric comparison for  (a) \(CD_t\) and (b) Quality: equally weighted average of \Cov{}, \(\cSP{}\). The categories are divided based on the complexity of the object structure.}
        \label{fig:per_cat_ecg_plot}
        \vspace{-8mm}
\end{figure}

\textbf{Per-Category Results:} \cref{fig:per_cat_ecg_plot} shows a quantitative comparison of performance metrics across various object categories for ECG\cite{pan_ecg_2020}, trained on CD (Chamfer Distance), EMD (Earth Mover's Distance), and \ourLoss{}. These object categories are grouped into three complexity levels: Easy, Medium, and Hard. This is to underscore performance based on object complexity type.

In \cref{fig:per_cat_ecg_plot}(a), the scores represent the L2 norm of the Chamfer Distance (\(cd_t\)). For instance, in the \textit{car} category, CD scores 4.34, EMD scores 2.98, and \ourLoss{} scores 2.34. This demonstrates that \ourLoss{} achieves a score that is significantly lower than CD and EMD, indicating its superior performance. Notably, for \textit{lamp}, \ourLoss{} scores 5.50 compared to CD's 17.88 and EMD's 6.20, and for \textit{sofa}, \ourLoss{} has a score of 3.68 against CD's 15.11 and EMD's 4.07. These scores indicate a consistent improvement in \(CD_t\) metric value if the model is trained with \ourLoss{}.

The plot in \cref{fig:per_cat_ecg_plot}(b) depicts Quality, computed as the average of the metrics: \Cov{},  \(\cSP{}\). Here, higher scores indicate better performance. Again, \ourLoss{} consistently attains the highest scores. For example, in \textit{car}, it scores 0.87, outperforming the next best scores by EMD (0.85), and while for \textit{chair} (\ourLoss{} 0.83) we can see a substantial improvement from CD (0.71) while also outperforming EMD (0.81). These results underscore the effectiveness of \ourLoss{} in not only minimizing \(cd_t\) but also in enhancing the overall quality of the models, as measured by Quality metrics.

Networks trained using \ourLoss{} produce point clouds with higher coverage and fewer spurious points compared to CD and EMD-based loss functions. This means that networks trained with \ourLoss{} are more effective at not only capturing local structure and point distribution but also in reducing SP that may cause the resulting shape to contain misaligned points and noise. This is particularly important in applications where the fidelity of the reconstructed shape to the true structure is critical.
Moreover, the quality score, which is an aggregate measure of the models' performance (\(COV,\cSP{}\)), is highest for models trained with \ourLoss{}. This captures the general superiority of \ourLoss{}-trained models over those trained with CD and EMD loss functions, confirming that \ourLoss{} is a superior loss function for training point cloud neural networks.
\subsection{Image to 3D Reconstruction}
\begin{figure*}[t]
  \vspace{-3em}
  \centering
  \includegraphics[width=0.8\textwidth]{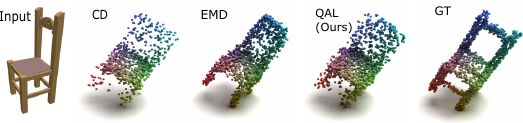}
  \vspace{-1.5em}
  \caption{Qualitative results from training PSG for 1000 epochs using CD, EMD, and \ourLoss{} loss.} % Left to Right: 2D Input, CD Loss, EMD Loss, \ourLoss{}, Ground Truth.}
  \label{fig:psg_visualization}
  \vspace{-1.5em}
\end{figure*}
Point Set Generation Network~\cite{fan2017point} is a pioneer in the single image to 3D reconstruction domain. We use PSG to evaluate the performance of \ourLoss{} by replacing CD and EMD loss used in PSG with \ourLoss{}. \cref{tab:P2M_net_with_pqm_loss} shows \ourLoss{} coverage improvements when compared with CD and EMD. As PSG is only trained for the 1000 epochs, the resolution doesn't increase, resulting in low resolution. Having said that, ~\cref{fig:intro} and \cref{fig:psg_visualization} show that qualitatively \ourLoss{} produces point clouds with high coverage and a much better appearance. Through these results, we show that Quality score, with the even weighting scheme, although meaningful for one application like \cref{subsec:compnet} is inadequate for this application. From the given table, we observe a drastic increase in the coverage when compared to CD and EMD. Since PSG considers a set of fixed-length random vectors as an input space, the resolution and \(\overline{SP}\) in the case of CD and EMD increases. However, the coverage metric is significantly low, causing a qualitative degradation where the object clearly appears skewed or sparse. 

\subsection{Image to Mesh}
A third application, Pixel2Mesh~\cite{Wang_2018_ECCV} is used to further scrutinize \ourLoss{}. Pixel2Mesh is a well-known benchmark for a single image to 3D mesh reconstruction, which uses CD loss with three other mesh-specific losses as their loss function. In the interest of consistency,
we swap CD loss with \ourLoss{} and use the same ShapeNet dataset as the authors.~\cref{tab:P2M_net_with_pqm_loss} shows that \ourLoss{} performs competitively.
%

% \begin{table*}[]
%     \centering
%     % \vspace{-0.2cm}
% %\begin{tabular}{lllllllllll}
%     \caption{Image to 3D points Generation and Mesh generation with \ourLoss{} Loss (Validation $\epsilon$=0.09m). PQM metrics are percentages. }
%     \vspace{-0.2cm}
%     \label{tab:P2M_net_with_pqm_loss}
% \begin{tabular}{@{}c|c|ccc||cccccc@{}}
% \toprule
% Network & Loss &  \(CD_t\) \(\downarrow\)  & \(EMD\) \(\downarrow\) &   \(F1 \uparrow\) & \(Cov.\) \(\uparrow\) &    \begin{tabular}[c]{@{}c@{}}\(Artifact\)\\ \(Score\) \(\uparrow\)\end{tabular} & \(Res.\) \(\uparrow\) & \(Acc.\) \(\uparrow\)  & \(PQM\) \(\uparrow\)    \\\midrule

% \multirow{3}{*}{P2M~\cite{Wang_2018_ECCV}} & CD & \textbf{0.0044} & 2726 & 80 & 78 & 77 & 72 & 83  & 78\\
% &QAL & 0.0049 & \textbf{1087} & \textbf{83} &\textbf{86} & \textbf{82} & \textbf{88} & \textbf{90}  & \textbf{86}\\
% \cmidrule{3-10}

% \multirow{3}{*}{PSG~\cite{fanPointSetGeneration2017}} & CD & 0.004 & 5586 & 68 & 49 & 48 & \textbf{98} & 68  & 50\\
%  & EMD &0.011  &3548 & 78 & 37 & 52 & 56 & 62  & 51 \\
%  & QAL & \textbf{0.002}& \textbf{2496} & \textbf{83} & \textbf{72}& \textbf{60} & 47& \textbf{69} & \textbf{62} \\
% \bottomrule
% \end{tabular}

% % \vspace{-1em}
% \end{table*}

\begin{table}[t]
    \centering
    % \vspace{-0.2cm}
%\begin{tabular}{lllllllllll}
    \caption{Image to 3D points Generation and Mesh generation with \ourLoss{} Loss (Validation $\epsilon$=0.09m, L1-CD $\times 1e^{3}$). Quality metrics are percentages. }
    \vspace{-0.2cm}
    \label{tab:P2M_net_with_pqm_loss}

\centering
\scriptsize
\setlength{\tabcolsep}{2pt}

\begin{tabular}{@{}c|c|ccc||cccc@{}}
\toprule
Network & Loss &  \(CD_t\) \(\downarrow\)  & \(EMD\) \(\downarrow\) &   \(F1 \uparrow\) & \(Cov.\) \(\uparrow\) &    \(\overline{SP}\) \(\uparrow\) & \(Quality\) \(\uparrow\)    \\\midrule

\multirow{3}{*}{P2M~\cite{Wang_2018_ECCV}} & CD & \textbf{0.0044} & 2726 & 0.80 & 78 & 77 & 78\\
&QAL & 0.0049 & \textbf{1087} & \textbf{0.83} &\textbf{86} & \textbf{82} & \textbf{86}\\
\cmidrule{3-8}

\multirow{3}{*}{PSG~\cite{fan2017point}} & CD & 0.004 & 5586 & 0.68 & 49 & 48 & 50\\
 & EMD &0.011  &3548 & 0.78 & 37 & 52  & 51 \\
 & QAL & \textbf{0.002}& \textbf{2496} & \textbf{0.83} & \textbf{72}& \textbf{60} & \textbf{62} \\
\bottomrule
\end{tabular}

\vspace{-0.3cm}
\end{table}

\begin{table}[t]
    \centering

    \caption{PCN with QAL on the lamp category at different output resolutions. 
QAL maintains strong coverage and quality across densities, indicating robustness to output point count even on a hard class with thin structures.(Validation $\epsilon$=0.03, L1-CD $\times 1e^{3}$)}
    \label{tab:resolution_table}
\vspace{-0.2cm}

\centering
\scriptsize
\setlength{\tabcolsep}{2pt}
\begin{tabular}{@{}c|c|cc||ccc@{}}
\toprule
Resolution & Loss &  \(CD_t\) \(\downarrow\)  &   \(F1 \uparrow\) & \(Cov.\) \(\uparrow\) & \(\overline{SP}\) \(\uparrow\)  & \(Quality\) \(\uparrow\)    \\
\midrule
\multirow[t]{3}{*}{2048} & EMD & 14.7 &  0.11 &  20.5   &  28.1  & 24.3 \\
& CD & 2.6 &  0.18      &  40.4   &  \textbf{46.0}  & 43.2 \\
 & QAL & \textbf{2.3} & \textbf{0.25}    &  \textbf{49.7}  &   45.5 & \textbf{47.6} \\
\cmidrule{1-7}
\multirow[t]{3}{*}{4096} &EMD & 15.9    &  0.14    &  20.5   &  29.0     & 27.3 \\
& CD  & 2.5   &  0.25    &  50.0   &  \textbf{47.0}  & 48.5 \\
& QAL & \textbf{2.1}   & \textbf{0.32}    &  \textbf{57.5}   &  46.5  & \textbf{52.0} \\
\cmidrule{1-7}
\multirow[t]{3}{*}{8192} & EMD & 18.0 & 0.17 & 28.0 & 34.3 & 31.1 \\
& CD   & 2.4 &  0.31   &  57.0  &  46.3   & 51.7 \\
& QAL  & \textbf{2.0} & \textbf{0.38}    &  \textbf{63.0}  &   \textbf{46.5} & \textbf{54.8} \\
% \bottomrule
% \cmidrule{1-7}
\bottomrule
\end{tabular}
\vspace{-0.6cm}
\end{table}

\subsection{Effect of Resolution}
\label{sec:resolution}

% In \cref{tab:resolution_table}, we evaluate training with \ourLoss{} at different target resolutions on PCN~\cite{yuan_pcn_2018} for 100 epochs. All other configurations are identical to prior experiments. Across resolutions, \ourLoss{} maintains strong performance with the same hyperparameters $(\epsilon,\omega,\lambda_{\mathrm{attr}})$, showing that resolution does not require retuning.

% Importantly, networks trained with \ourLoss{} consistently produce outputs with higher coverage and lower or comparable spurious point (SP) than CD/EMD, even at coarse resolutions. This indicates that the coverage–attraction design of QAL generalizes robustly to varying output densities, without sacrificing quality or stability.

\cref{tab:resolution_table} reports PCN trained with QAL at different output resolutions (2048, 4096, 8192 points), using the \textbf{lamp category} as it is considered one of the most challenging classes due to its thin structures and fine details. 
Performance remains consistent across densities, showing that QAL generalizes robustly to varying output point counts without requiring re-tuning.

\subsection{Impact of Completion on Grasp Utility}
\label{sec:grasp_utility}
Beyond geometric metrics, we investigate how improved completion and coverage affect the utility of reconstructed point clouds for manipulation. We evaluate ECG models trained on the MVP dataset with all the losses considered in \cref{tab:qal_pcn_evaluation}. To focus on the effect of high-coverage completions, we sample the top 1500 QAL reconstructions ranked by Cov@$\epsilon$ and evaluate them, along with corresponding predictions from other baselines, using a pre-trained GraspNet model~\cite{fang2020graspnet}. The model predicts candidate grasps and assigns a grasp quality score, providing a task-driven evaluation: if completions better preserve object geometry, they should yield higher grasp scores and enable more feasible grasps.

\cref{tab:grasp_score_table} reports the results across six losses. QAL achieves the \textbf{highest average grasp score (0.5417 $\pm$ 0.3461)}, while EMD yields the largest number of feasible grasps (108.8 on average). These results highlight a trade-off: coverage-oriented objectives like QAL lead to higher-quality grasps, whereas EMD emphasizes grasp diversity. Overall, the findings validate that completion quality and coverage directly enhance the downstream utility of point clouds for grasp manipulation.

\begin{table}[t]
    \centering
    \caption{ECG model predictions evaluated for grasp score and number of grasps across different loss functions on the GraspNet-1Billion dataset~\cite{fang2020graspnet}. Best mean score is highlighted in bold.}
    \label{tab:grasp_score_table}
    \vspace{-0.2cm}

    \scriptsize
    \setlength{\tabcolsep}{6pt}
    \begin{tabular}{@{}c|cc@{}}
    \toprule
    Loss & Grasp Score (Mean $\pm$ Std) & Num Grasps \\
    \midrule
    CD       & 0.5379 $\pm$ 0.3456 & 85.54 \\% $\pm$ 162.60 \\
    EMD      & 0.5239 $\pm$ 0.3948 & 108.77 \\%$\pm$ 194.62 \\

    HCD      & 0.5264 $\pm$ 0.3444 & 82.19 \\%$\pm$ 158.58 \\
    DCD      & 0.5324 $\pm$ 0.3484 & 87.98 \\%$\pm$ 165.07 \\
    INFOCD   & 0.5353 $\pm$ 0.3478 & 86.10 \\%$\pm$ 163.55 \\
    \textbf{QAL}      & \textbf{0.5417 $\pm$ 0.3461} & 90.48 \\%$\pm$ 168.66 \\
    \bottomrule
    \end{tabular}
    \vspace{-0.5cm}
\end{table}

% \section{Discussion}

% \noindent\textbf{What QAL optimizes.} 
% QAL makes the recall–precision trade-off explicit: its coverage-weighted term emphasizes errors beyond a tolerance $\epsilon$ (precision proxy), while the uncovered--GT attraction activates where recall is missing. This aligns training with thresholded metrics (Cov, SP, F1) commonly used in reconstruction and robotics \cite{Tatarchenko_2019_CVPR,varley2017shape,lundell2020beyond}. Across backbones, QAL consistently improves Coverage, with its effect on SP controlled by $\lambda_{\mathrm{attr}}$.

% \noindent\textbf{Quantitative vs. qualitative impact.} 
% Although average Coverage gains over CD and InfoCD are modest (+3--4 pts), they yield disproportionately large qualitative benefits. As shown in \cref{fig:pcn_visualization}, even slight increases in coverage recover sharper edges and fine structures that other losses miss—critical for manipulation and navigation.

% \noindent\textbf{Practicality and limitations.} 
% Ablations indicate stable defaults ($\epsilon{=}10^{-3}$, $\omega{=}10$, $\lambda_{\mathrm{attr}}{\in}[0.5,1.0]$), though tuning may be needed depending on tolerance for false positives. 
% Future work includes adaptive scheduling of $\epsilon$ or $\lambda_{\mathrm{attr}}$, and extending QAL beyond point sets to robotics tasks such as grasp planning and safe navigation. 
% QAL performs robustly across diverse settings, though highly occluded or ambiguous shapes remain difficult (see \cref{fig:suppl_bad} in the supplementary material), suggesting that stronger shape priors may further improve performance.

\section{Discussion}

\noindent\textbf{What QAL optimizes.} 
QAL makes the recall–precision trade-off explicit: its coverage-weighted term emphasizes errors beyond a tolerance $\epsilon$ (precision proxy), while the uncovered--GT attraction activates where recall is missing. 
This aligns training with thresholded metrics such as Coverage, Spurious Points, and F1, providing a more diagnostic view of performance than CD or EMD alone. 
Across backbones, QAL consistently improves Coverage, with the effect on spurious points moderated by $\lambda_{\mathrm{attr}}$.

% \noindent\textbf{Quantitative vs. qualitative impact.} 
% Although average Coverage gains over CD and its variants are modest (+3--4 pts), they correspond to meaningful improvements in geometry. 
% As illustrated in \cref{fig:pcn_visualization}, even small increases in coverage restore thin structures and sharper edges that CD-based models typically miss. 
% These qualitative differences explain why QAL-trained completions yield higher grasp scores in GraspNet evaluation: improved completeness provides more reliable surfaces for downstream use.
\noindent\textbf{Quantitative vs. qualitative impact.} 
Although average Coverage gains over CD and its variants appear modest (+3--4 pts), they correspond to meaningful improvements in geometry. 
As illustrated in \cref{fig:pcn_visualization}, even small increases in coverage restore thin structures and sharper edges that CD-based models typically miss. 
CD and F1, while standard, often provide ambiguous signals—CD may differ even when recall/precision are unchanged, and F1 can remain flat despite recovered details—whereas \Cov{} and $\overline{\mathrm{SP}}$ capture these improvements more faithfully. 
Further evaluation in the supplementary (see \cref{sec:effect_of_loss_on_ae}) confirms this trend, showing that QAL better preserves detail and suppresses artifacts even in simple reconstruction settings. 
This explains why QAL-trained completions also yield higher grasp scores in GraspNet evaluation: improved completeness provides more reliable surfaces for downstream use.

\noindent\textbf{Practicality and limitations.} 
Ablations show that QAL performs robustly across a range of hyperparameters and output resolutions, with $\epsilon{=}10^{-3}$, $\omega{=}10$, and $\lambda_{\mathrm{attr}}{\in}[0.5,1.0]$ serving as stable defaults. 
Nonetheless, highly occluded or ambiguous shapes remain challenging, where the attraction term may introduce mild over-prediction (see \cref{fig:suppl_bad} in the supplementary material). 
These cases suggest future directions such as incorporating stronger shape priors or adaptive scheduling of $\epsilon$ and $\lambda_{\mathrm{attr}}$ to further refine the recall–precision balance. Finally, our grasp evaluation relies on GraspNet’s simulated scoring rather than physical grasp trials, so confirming these gains on real hardware remains future work.

\section{Conclusion}
We introduced \textbf{QAL}, a differentiable loss that decouples coverage and attraction terms to explicitly balance recall and precision. 
Across four completion backbones, QAL improves Coverage by an average of \textbf{+4.3 pts}, consistently recovering thin structures and details missed by CD- and EMD-based models. 
We further demonstrate generalization beyond point cloud completion, showing consistent improvements in single-image 3D reconstruction and image-to-mesh generation, as well as higher grasp scores under GraspNet evaluation. 
As a simple, interpretable drop-in replacement wherever CD/EMD or their variants are used, QAL provides a principled way to align training with thresholded metrics such as Coverage and F1. 
Its recall-oriented formulation makes it broadly applicable across 3D vision pipelines, with potential benefits for large-scale reconstruction and safety-critical domains where missing geometry is particularly costly.

{
    \small
    \bibliographystyle{ieeenat_fullname}
    \bibliography{pointcloudlidar}
}

\clearpage

\renewcommand{\thesection}{\Alph{section}}
% \section{QAL: Supplementary Material}

% We provide extended qualitative results of the ECG~\cite{pan_ecg_2020} baseline trained on the \textbf{MVP} dataset~\cite{pan_variational_2021}. Figure~\ref{fig:suppl} illustrates completions across a diverse set of object categories. In several cases, ECG produces visually plausible completions that recover coarse geometry and overall object symmetry, particularly for relatively simple or well-observed shapes. However, we also observe consistent failure modes: thin structures (e.g., table legs, lamp stands) are often missing, flat surfaces may appear warped, and spurious points emerge in regions with heavy self-occlusion. These artifacts manifest as low recall and low precision, respectively, and highlight the sensitivity of CD-based training to complex geometric features. The examples further reinforce our motivation: even a strong baseline like ECG fails to reliably capture fine details, motivating the need for losses such as QAL that explicitly regulate the recall–precision trade-off.

\section{QAL: Supplementary Material}

\begin{figure*}[!ht]
  \centering
  \vspace{-0.4cm}
  \begin{tabular}{c}
    \includegraphics[width=\textwidth]{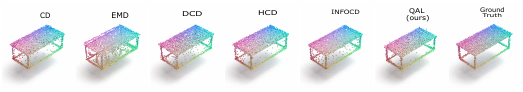} \\[-0.4cm]
    \includegraphics[width=\textwidth,trim=0 0 0 1.2cm,clip=true]{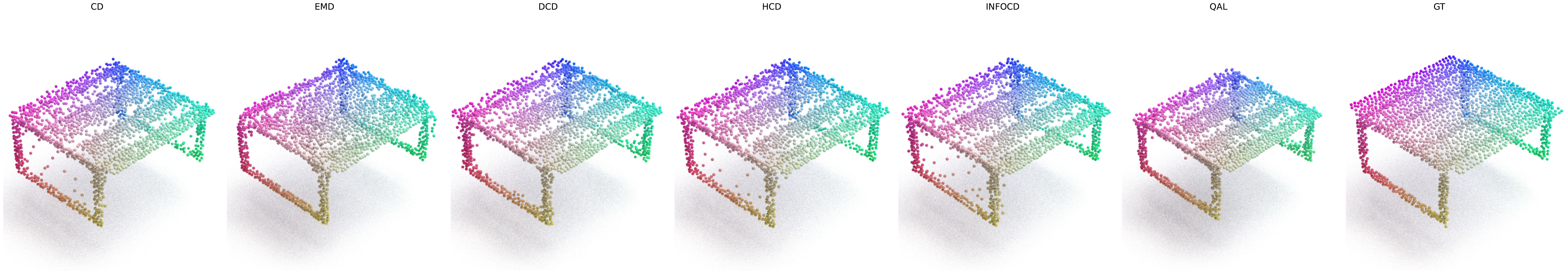} \\[-0.4cm]
    \includegraphics[width=\textwidth,trim=0 0 0 1.2cm,clip=true]{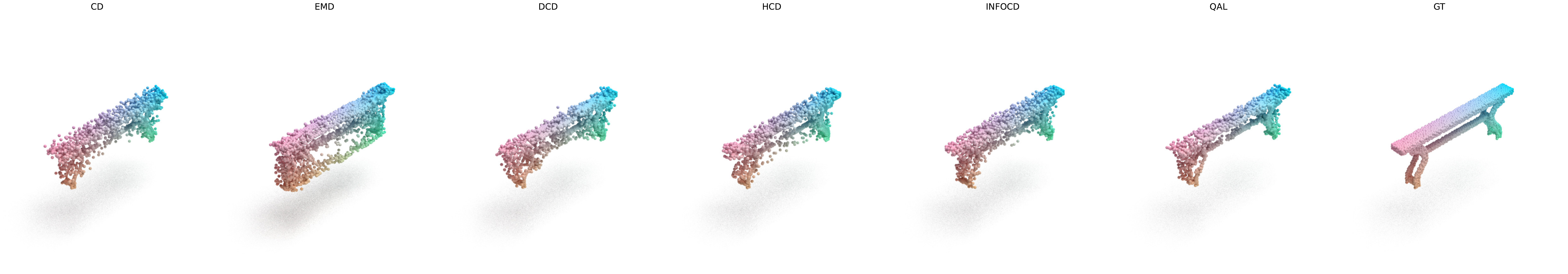} \\[-0.4cm]
    \includegraphics[width=\textwidth,trim=0 0 0 1.2cm,clip=true]{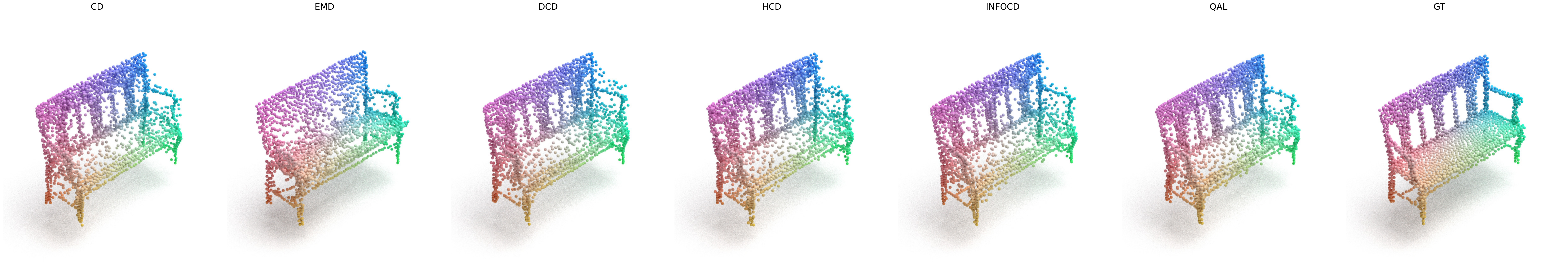} \\[-0.4cm]
    \includegraphics[width=\textwidth,trim=0 0 0 1.2cm,clip=true]{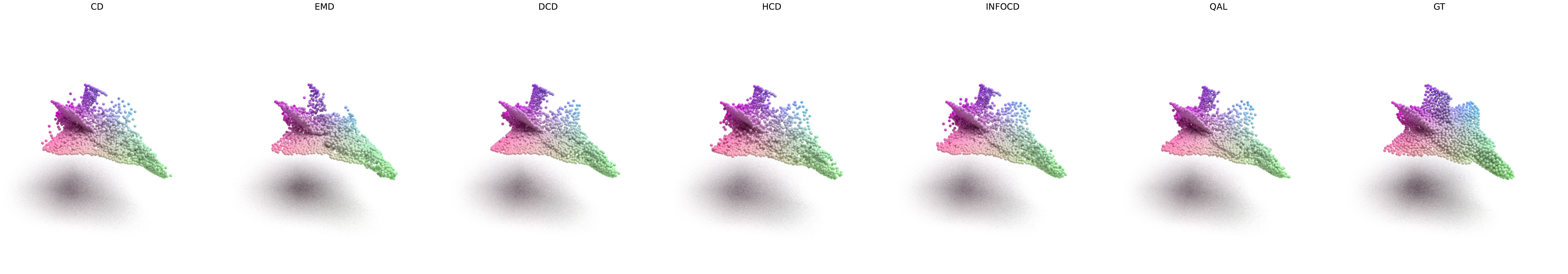} \\[-0.4cm]

    \includegraphics[width=\textwidth,trim=0 0 0 1.2cm,clip=true]{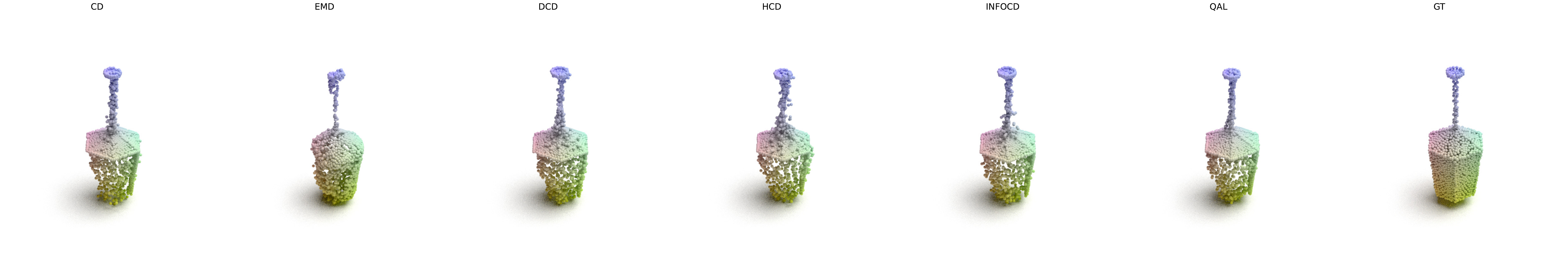} \\[-0.4cm]
    \includegraphics[width=\textwidth,trim=0 0 0 1.2cm,clip=true]{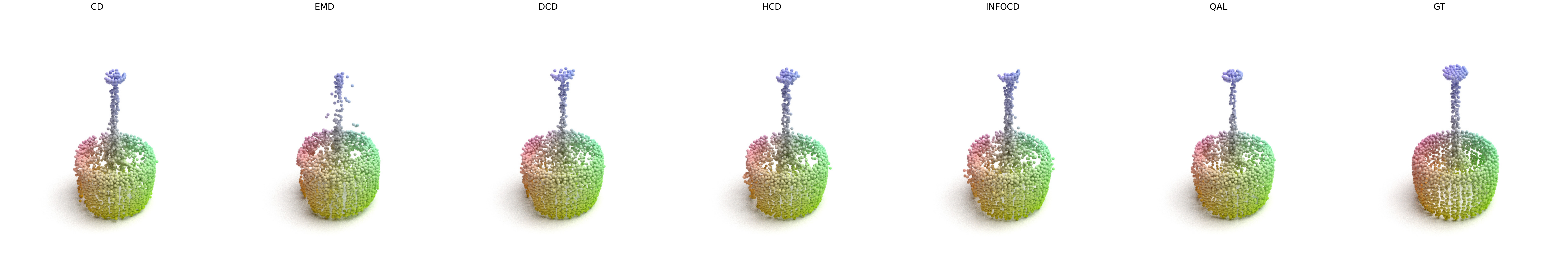} \\[-0.4cm]
    \includegraphics[width=\textwidth,trim=0 0 0 1.2cm,clip=true]{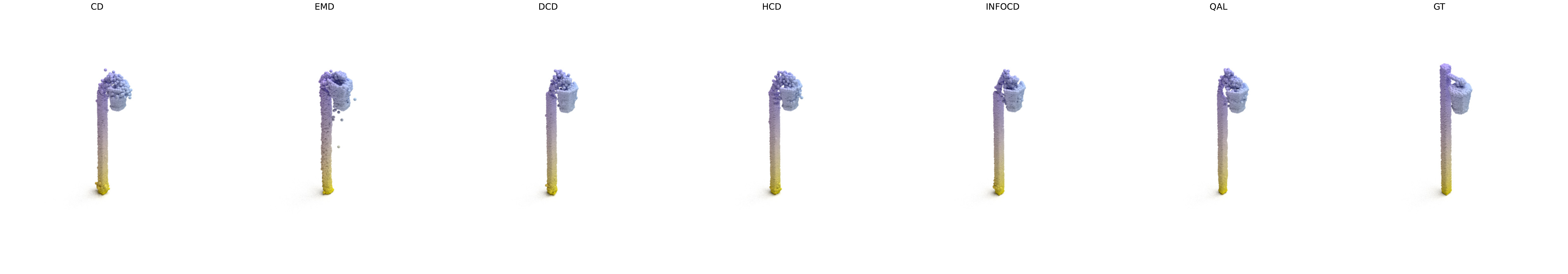} \\[-0.4cm]
  \end{tabular}
  \vspace{-0.6cm}
  \caption{Qualitative comparisons of point cloud completion across 8 representative object categories from the \textbf{MVP} dataset~\cite{pan2021variational}. Each row shows an input partial point cloud and completions generated by ECG~\cite{pan_ecg_2020} trained with CD, EMD, DCD, HCD, InfoCD, and our proposed QAL, followed by the ground truth. QAL consistently recovers thin structures (e.g., chair legs, lamp posts) and suppresses spurious artifacts, producing reconstructions that are both more complete and geometrically faithful compared to existing losses. Input partial clouds are omitted for space; all methods use identical inputs.}
  \label{fig:suppl_good}
  \vspace{-0.4cm}
\end{figure*}

We provide extended qualitative comparisons on the \textbf{MVP} dataset~\cite{pan2021variational} using the ECG~\cite{pan_ecg_2020} backbone trained with different loss functions. Figure~\ref{fig:suppl_good} shows representative success cases. Across diverse object categories, QAL yields completions with \emph{fewer spurious points}, better preservation of \emph{thin structures} such as chair legs and lamp posts, and sharper recovery of \emph{fine geometric details} compared to CD, EMD, DCD, HCD, and InfoCD. These improvements demonstrate QAL’s ability to balance recall and precision, producing reconstructions that are both complete and clean.

\begin{figure*}[!b]
  \centering
  % \vspace{-0.4cm}
  \begin{tabular}{c}
    \includegraphics[width=\textwidth]{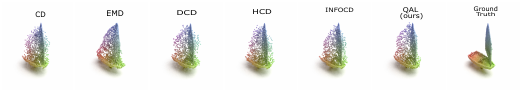} \\[-0.2cm]
    \includegraphics[width=\textwidth,trim=0 0 0 1.2cm,clip=true]{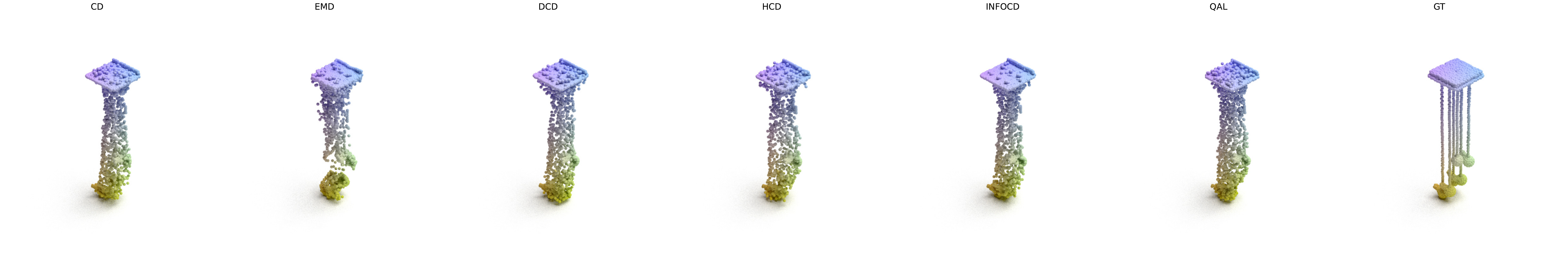} \\[-0.2cm]
  \end{tabular}
  \vspace{-0.4cm}
  \caption{Failure cases of point cloud completion on challenging \textbf{MVP} samples~\cite{pan2021variational}. Each row shows an input partial point cloud and completions generated by ECG~\cite{pan_ecg_2020} trained with CD, EMD, DCD, HCD, InfoCD, and our proposed QAL, followed by the ground truth. While QAL generally outperforms existing losses, certain instances with complex geometry remain difficult: thin or elongated parts may be under-reconstructed, and spurious points can still appear around highly ambiguous regions. These examples highlight open challenges in reliably capturing fine details under severe incompleteness. Input partial clouds are omitted for space; all methods use identical inputs.}
  \label{fig:suppl_bad}
  \vspace{-0.4cm}
\end{figure*}

To complement these strengths, Figure~\ref{fig:suppl_bad} presents challenging failure cases. While QAL generally outperforms existing losses, objects with highly irregular or ambiguous geometry remain difficult: elongated parts may still be under-reconstructed, flat surfaces can appear distorted, and spurious points may emerge under severe occlusion. These examples highlight open challenges in reliably capturing fine details from severely incomplete observations and suggest promising directions for future work on stronger priors and shape reasoning.

\section{Per-Category Coverage Analysis}

To provide a more detailed view of reconstruction quality, we report \textbf{per-category coverage} across all models and loss functions in Table~\ref{tab:coverage_per_category}. 
While the main paper highlights average performance, these results reveal finer trends within individual object categories of the MVP dataset. 

Several observations emerge. First, \textbf{QAL consistently achieves the highest coverage across most categories and models}, often improving upon CD, HCD, and InfoCD. For instance, on \textit{airplane}, \textit{cabinet}, and \textit{table} objects, QAL provides clear gains over both CD and InfoCD. Second, although EMD enforces one-to-one correspondences, its coverage remains markedly lower across categories, underscoring its tendency to produce over-smoothed reconstructions. Finally, categories with thin or fine-grained structures such as \textit{lamp} and \textit{chair} show the greatest benefit from QAL, where balancing recall and precision is critical to capturing delicate geometric details.

These category-level results complement the aggregate analysis and further validate the robustness of QAL: it improves completeness of reconstructions without sacrificing precision, even on challenging object classes.

\section{Evaluation Protocol and Trade-off Analysis}
\noindent\textbf{Choice of $\epsilon$ for Coverage Evaluation}
We report coverage at a fixed threshold of $\epsilon=0.03$. 
This value was selected based on the average nearest-neighbor spacing in MVP samples at 2048 points, which is approximately $0.0156$. 
Setting $\epsilon$ to twice this mean spacing provides a strict enough tolerance to penalize missing thin structures, while remaining flexible enough to ignore minor surface noise or small alignment discrepancies. 
Empirically, thresholds lower than $0.02$ excessively penalized completions for surface noise, while thresholds above $0.05$ blurred differences between methods.

\noindent\textbf{Reporting Trade-offs}
Recall-oriented training with QAL naturally induces a trade-off between recall and precision. 
To highlight this, we present grouped bar plots in our ablations, showing \(\mathrm{Cov}\) ($\uparrow$), the complement of spurious points \(\overline{\mathrm{SP}}\) ($\uparrow$), and Chamfer Distance (CD; $\downarrow$) side-by-side for each hyperparameter setting. 
This visualization makes trade-offs explicit without collapsing into a single score. 
While we report an aggregate quality metric in the main paper for completeness, we emphasize that per-metric reporting provides clearer insight into how QAL balances recall and precision.

\begin{table}[t]
\centering
\scriptsize
\setlength{\tabcolsep}{1.5pt}
\caption{Per-category coverage (\%) across models and loss functions on the MVP dataset. QAL consistently achieves the highest coverage, with best results highlighted in bold.(Validation $\epsilon$=0.03)}
\label{tab:coverage_per_category}
    \begin{tabular}{llrrrrrrrrr}
    \toprule
    Model & Loss & Airplane & Cabinet & Car & Chair & Lamp & Sofa & Table & Boat & Average \\

\midrule
    \multirow{6}{*}{\textbf{PCN}} & \textbf{CD} & 77.0 & 55.3 & 62.3 & 50.9 & 56.1 & 53.3 & 56.4 & 67.1 & 59.8 \\
 & \textbf{EMD} & 34.5 & 13.2 & 20.8 & 17.2 & 11.0 & 27.0 & 30.7 & 14.2 & 21.1 \\
 & \textbf{DCD} & 78.5 & \textbf{61.2} & 67.0 & 54.0 & 56.6 & 58.7 & 59.6 & 70.4 & 63.2 \\
 & \textbf{HCD} & 77.0 & 56.0 & 63.1 & 50.3 & 54.3 & 53.3 & 57.3 & 67.3 & 59.8 \\
 & \textbf{InfoCD} & 77.6 & 58.2 & 65.9 & 55.1 & 59.3 & 57.1 & 58.9 & 70.1 & 62.8 \\
 & \textbf{QAL} & \textbf{79.7} & 61.1 & \textbf{68.4} & \textbf{58.1} & \textbf{62.2} & \textbf{60.7} & \textbf{62.6} & \textbf{74.5} & \textbf{65.9} \\
\cmidrule{2-11}
\multirow{4}{*}{\textbf{TOPNET}} & \textbf{CD} & 75.0 & 49.7 & 57.0 & 45.6 & 49.8 & 47.1 & 52.0 & 65.7 & 55.2 \\
 % & \textbf{EMD} & -- & -- & -- & -- & -- & -- & -- & -- & -- \\
 % & \textbf{DCD} & -- & -- & -- & -- & -- & -- & -- & -- & -- \\
 & \textbf{HCD} & 75.5 & 48.7 & 56.0 & 45.9 & 53.0 & 47.8 & 52.4 & 67.0 & 55.8 \\
 & \textbf{InfoCD} & 74.9 & 49.7 & 57.9 & 46.0 & 55.2 & 47.4 & 52.8 & 66.2 & 56.3 \\
 & \textbf{QAL} & \textbf{77.7} & \textbf{51.1} & \textbf{60.0} & \textbf{49.5} & \textbf{57.9} & \textbf{50.7} & \textbf{55.3} & \textbf{70.4} & \textbf{59.1} \\
\cmidrule{2-11}
\multirow{6}{*}{\textbf{ECG}} & \textbf{CD} & 85.7 & 56.0 & 60.2 & 61.4 & 71.5 & 56.7 & 68.0 & 73.2 & 66.6 \\
 & \textbf{EMD} & 83.1 & \textbf{59.5} & 61.5 & 61.1 & 71.2 & \textbf{59.8} & 66.9 & 72.8 & 67.0 \\
 & \textbf{DCD} & 86.0 & 58.6 & 62.8 & 62.7 & 72.1 & 58.6 & 69.8 & 74.5 & 68.1 \\
 & \textbf{HCD} & 85.4 & 57.0 & 60.3 & 61.7 & 71.5 & 56.9 & 68.7 & 72.8 & 66.8 \\
 & \textbf{InfoCD} & 85.3 & 57.0 & 61.8 & 62.1 & 71.3 & 57.7 & 66.7 & 74.0 & 67.0 \\
 & \textbf{QAL} & \textbf{87.2} & 59.3 & \textbf{63.1} & \textbf{63.4} & \textbf{73.3} & 59.2 & \textbf{70.0} & \textbf{75.8} & \textbf{68.9} \\
\cmidrule{2-11}
\multirow{3}{*}{\textbf{Seedformer}} & \textbf{CD} & 86.1 & 67.4 & 66.4 & 69.4 & 77.8 & 66.6 & 72.9 & 77.7 & 73.0 \\
 % & \textbf{EMD} & -- & -- & -- & -- & -- & -- & -- & -- & -- \\
 % & \textbf{DCD} & -- & -- & -- & -- & -- & -- & -- & -- & -- \\
 % & \textbf{HCD} & -- & -- & -- & -- & -- & -- & -- & -- & -- \\
 & \textbf{InfoCD} & 76.7 & 38.4 & 40.1 & 50.7 & 64.0 & 41.5 & 56.9 & 58.8 & 53.4 \\
 & \textbf{QAL} & \textbf{89.5} & \textbf{70.0} & \textbf{73.5} & \textbf{73.8} & \textbf{81.1} & \textbf{72.4} & \textbf{77.4} & \textbf{85.1} & \textbf{77.8} \\
% \cmidrule{2-11}
    \bottomrule
    \end{tabular}
    \end{table}

\section{Effect of Loss on Reconstruction.} 
\label{sec:effect_of_loss_on_ae}
Using the same trained models from Section~\ref{subsec:compnet} (Table~\ref{tab:qal_pcn_evaluation}), 
we further evaluate reconstruction performance when ground-truth point clouds are provided as input on the MVP dataset (1200 samples). 
This controlled setting removes the challenge of inferring missing regions and highlights how the choice of loss function alone shapes the fidelity of the reconstructed geometry. 
Table~\ref{tab:qal_ae_eval} shows that while CD and F1 remain standard, they often provide incomplete or even misleading signals. 
For example, in PCN, QAL improves Coverage by +4.5 pts over CD (65.3 vs. 60.8) while CD and F1 remain nearly unchanged, masking the recovery of fine details. 
In TOPNet, CD and F1 suggest negligible differences across losses, yet Coverage and Quality reveal that QAL yields denser and more uniform reconstructions (+4.3 pts Coverage over CD). 
In ECG, HCD and QAL achieve near-identical \Cov{} (91.6) and $\overline{\mathrm{SP}}$ ($\sim$81), yet CD differs (6.0 vs. 5.8), illustrating that CD alone can produce misleading signals even when recall and precision are effectively the same. 
Finally, in SeedFormer, F1 stays nearly flat across CD, InfoCD, and QAL (0.62–0.66), while Coverage rises from 87.8 (CD) to 88.3 (QAL), and Quality remains competitive despite InfoCD reporting the lowest CD. 
% This shows that CD and F1 alone cannot capture the nuanced trade-offs between recall and precision.

% These cases highlight that CD favors small average distances, while F1 is threshold-sensitive and may improve even if reconstructions cluster unnaturally, leading to contradictory interpretations. 
% By contrast, the additional dimensions introduced here—\Cov{} and $\overline{\mathrm{SP}}$—capture complementary geometric aspects: whether fine details are covered and whether spurious points are suppressed. 
% Together, they resolve the ambiguity of CD/F1 and provide a more faithful picture of reconstruction quality, confirming that \ourLoss{} not only benefits partial-to-complete completion but also enhances detail preservation in autoencoding scenarios.

\begin{figure*}[!ht]
  \centering
  \vspace{-0.4cm}
  % \begin{tabular}{c}
  %   \includegraphics[width=\textwidth]{images/ae/pcn/1.pdf} \\[-0.4cm]
  %   \includegraphics[width=\textwidth,trim=0 0 0 1.2cm,clip=true]{images/ae/pcn/208.pdf} \\[-0.4cm]
  %   \includegraphics[width=\textwidth,trim=0 0 0 1.2cm,clip=true]{images/ae/pcn/238.pdf} \\[-0.4cm]
  % \end{tabular}
  \includegraphics[width=\textwidth]{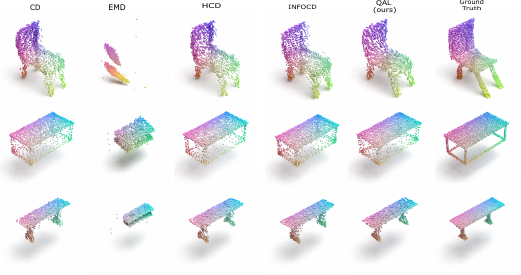}
  \vspace{-0.8cm}
  \caption{Representative reconstructions with PCN.  
CD and HCD reconstructions leave gaps in thin structures, while EMD collapses coverage with poor fidelity.  
InfoCD improves coverage modestly but introduces spurious clusters.  
QAL (ours) yields denser and more uniform point sets (+4.5 pts Cov.\ over CD), visibly recovering sharper chair backs and object contours. Input partial clouds are omitted for space; all methods use identical inputs.}

  \label{fig:ae_pcn}
  % \vspace{-0.4cm}
\end{figure*}

\begin{figure*}[!ht]
  \centering
  % \vspace{-0.4cm}
  % \begin{tabular}{c}
  %   \includegraphics[width=\textwidth]{images/ae/topnet/1.pdf} \\[-0.4cm]
  %   \includegraphics[width=\textwidth,trim=0 0 0 1.2cm,clip=true]{images/ae/topnet/238.pdf} \\[-0.4cm]
  %   \includegraphics[width=\textwidth,trim=0 0 0 1.2cm,clip=true]{images/ae/topnet/717.pdf} \\[-0.4cm]
  % \end{tabular}
  \includegraphics[width=\textwidth]{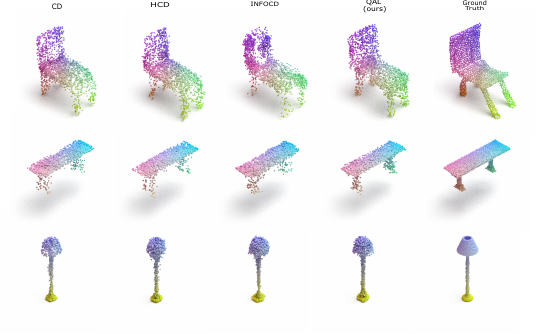}
  \vspace{-1.2cm}
  \caption{Representative reconstructions with TOPNet.  
CD, HCD, and InfoCD produce completions that look visually similar, consistent with flat CD/F1 values.  
However, QAL improves coverage (+4.3 pts Cov.\ over CD), filling in missing areas and producing more uniform reconstructions.  
These examples illustrate how CD/F1 miss improvements that are evident with coverage-aware evaluation. Input partial clouds are omitted for space; all methods use identical inputs.}

  \label{fig:ae_topnet}
  % \vspace{-0.4cm}
\end{figure*}

\begin{figure*}[!ht]
  \centering
  \vspace{-0.4cm}
  % \begin{tabular}{c}
  %   \includegraphics[width=\textwidth]{images/ae/ecg/1.pdf} \\[-0.4cm]
  %   \includegraphics[width=\textwidth,trim=0 0 0 1.2cm,clip=true]{images/ae/ecg/167.pdf} \\[-0.4cm]
  %   \includegraphics[width=\textwidth,trim=0 0 0 1.2cm,clip=true]{images/ae/ecg/208.pdf} \\[-0.4cm]
  %   \includegraphics[width=\textwidth,trim=0 0 0 1.2cm,clip=true]{images/ae/ecg/249.pdf} \\[-0.4cm]
  % \end{tabular}
  \includegraphics[width=0.9\textwidth]{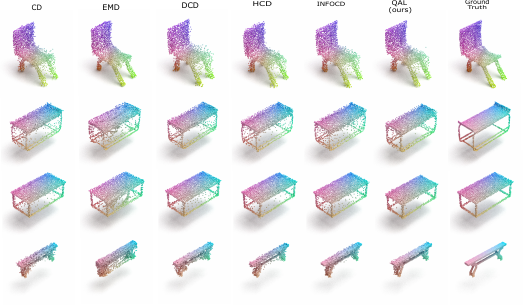}
  \vspace{-1.0cm}
   \caption{Representative reconstructions with ECG.  
    CD and DCD achieve good overall alignment but leave subtle missing details.  
    HCD and QAL both reach high coverage (91.6), yet CD differs (6.0 vs.\ 5.8), showing how CD alone can give misleading signals.  
    Qualitatively, QAL better balances recall and precision, producing sharper and more consistent reconstructions without adding spurious points.}
    
  \label{fig:ae_ecg}
  \vspace{-0.4cm}
\end{figure*}

\begin{figure*}[!ht]
  \centering
  \vspace{-0.4cm}
  % \begin{tabular}{c}
  %   \includegraphics[width=\textwidth]{images/ae/seedformer/30.pdf} \\[-0.4cm]
  %   \includegraphics[width=\textwidth,trim=0 0 0 1.2cm,clip=true]{images/ae/seedformer/238.pdf} \\[-0.4cm]
  %   \includegraphics[width=\textwidth,trim=0 0 0 1.2cm,clip=true]{images/ae/seedformer/626.pdf} \\[-0.4cm]
  %   \includegraphics[width=\textwidth,trim=0 0 0 1.2cm,clip=true]{images/ae/seedformer/1194.pdf} \\[-0.4cm]
  % \end{tabular}
  \includegraphics[width=0.6\textwidth]{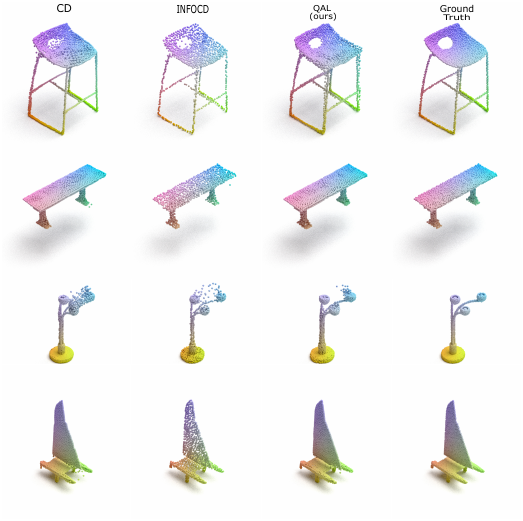}

  \vspace{-0.8cm}
    \caption{Representative reconstructions with SeedFormer.  
    CD reconstructions miss fine chair details and introduce spurious points around thin lamp structures, while InfoCD collapses coverage (75.1) and clusters points densely in certain regions, leaving other areas sparse.  
    In contrast, QAL restores coverage to 88.3 while maintaining balance between detail recovery and noise suppression, yielding more faithful reconstructions across both thin and coarse structures. Input partial clouds are omitted for space; all methods use identical inputs.}

  \label{fig:ae_seedformer}
  % \vspace{-0.4cm}
\end{figure*}

This confirms that CD and F1 alone cannot capture the nuanced trade-offs between recall and precision.
% To complement these quantitative results, we also include 3–4 representative visualizations per model in the supplementary material. 
To complement these quantitative results, we also include 3–4 representative visualizations per model (\cref{fig:ae_pcn,fig:ae_topnet,fig:ae_ecg,fig:ae_seedformer}).
These examples qualitatively confirm the same trends: QAL consistently recovers sharper structures and reduces holes relative to CD/EMD, while avoiding the spurious clusters sometimes induced by InfoCD or HCD. 
Such visualizations reinforce that the geometric advantages highlighted by \Cov{} and $\overline{\mathrm{SP}}$ are clearly visible in the reconstructions.

These cases highlight that CD favors small average distances, while F1 is threshold-sensitive and may improve even if reconstructions cluster unnaturally, leading to contradictory interpretations. 
By contrast, the additional dimensions introduced here—\Cov{} and $\overline{\mathrm{SP}}$—capture complementary geometric aspects: whether fine details are covered and whether spurious points are suppressed. 
Together, they resolve the ambiguity of CD/F1 and provide a more faithful picture of reconstruction quality, confirming that \ourLoss{} not only benefits partial-to-complete completion but also enhances detail preservation in autoencoding scenarios.

\begin{table}[t]
\centering
\caption{Completion Networks trained with CD (L2-CD $\times 1e^{3}$), HCD, InfoCD and \textbf{QAL} for $\epsilon =0.03$. 
\Cov{}, $\overline{SP}$ and Quality metrics are scaled as percentages.}
\vspace{-0.2cm}
\scriptsize
\setlength{\tabcolsep}{4pt}
\label{tab:qal_ae_eval}
\begin{tabular}{l|cc|ccc}
\toprule
Method (Network+Loss) & CD ($\downarrow$) & F1 ($\uparrow$) & Cov. ($\uparrow$) & $\overline{SP}$ ($\uparrow$) & Quality ($\uparrow$) \\
\midrule
PCN + CD    & 17.0 & 0.30 & 60.8 & \textbf{64.6} & 62.7 \\
PCN + EMD   & 44.0 & 0.11 & 21.9 & 41.3 & 31.6 \\
PCN + HCD   & 17.2 & \textbf{0.31} & 60.4 & 64.5 & 62.4 \\
PCN + INFO  & \textbf{16.9} & 0.30 & 63.6 & 62.7 & 63.2 \\
PCN + QAL   & 16.9 & \textbf{0.31} & \textbf{64.8} & 63.1 & \textbf{64.0} \\
\cmidrule{1-6}
TOPNET + CD   & \textbf{17.7} & \textbf{0.28} & 56.2 & \textbf{60.9} & 58.6 \\
TOPNET + HCD  & 17.8 & \textbf{0.28} & 56.8 & 60.1 & 58.4 \\
TOPNET + INFO & \textbf{17.7} & \textbf{0.28} & 57.4 & 60.3 & 58.8 \\
TOPNET + QAL  & 17.8 & 0.27 & \textbf{60.5} & 57.9 & \textbf{59.2} \\
\cmidrule{1-6}
ECG + CD    & 6.2 & \textbf{0.82} & 90.9 & 80.6 & 85.8 \\
ECG + EMD   & 11.8 & 0.49 & 82.8 & 77.3 & 80.0 \\
ECG + DCD   & 6.1 & 0.80 & 91.0 & 79.9 & 85.4 \\
ECG + HCD   & 6.0 & \textbf{0.82} & \textbf{91.6} & \textbf{81.0} & \textbf{86.3} \\
ECG + INFO  & 6.6 & 0.81 & 89.6 & 79.7 & 84.6 \\
ECG + QAL   & \textbf{5.8} & 0.81 & \textbf{91.6} & 80.9 & 86.2 \\
\cmidrule{1-6}
SEEDFORMER + CD   & 9.4 & 0.62 & 87.8 & 87.8 & \textbf{87.8} \\
SEEDFORMER + INFO & \textbf{5.9} & 0.65 & 75.1 & \textbf{95.8} & 85.4 \\
SEEDFORMER + QAL  & 8.4 & \textbf{0.66} & \textbf{88.3} & 87.1 & 87.7 \\
\bottomrule
\end{tabular}
\vspace{-0.3cm}
\end{table}

\end{document}